\documentclass[review]{elsarticle}
\usepackage{natbib}
\usepackage[margin=1in]{geometry}
\usepackage[table]{xcolor}

\usepackage{lineno,hyperref}
\modulolinenumbers[5]

\journal{Journal of Biomedical Informatics}

\DeclareUnicodeCharacter{2212}{-}

\usepackage{algorithm}
\usepackage{algorithmic}
\usepackage{amsmath,amssymb,amsfonts}
\usepackage{booktabs}
\usepackage{caption}
\usepackage{subcaption}
\usepackage{cuted}
\usepackage{cleveref}
\usepackage{graphicx}
\usepackage{lipsum}
\usepackage{mathtools}
\usepackage{setspace}
\usepackage{textcomp}
\usepackage{tabularx}
\usepackage{url}
\usepackage[flushleft]{threeparttable}
\usepackage{xcolor}
\usepackage[normalem]{ulem}
\usepackage{multirow}
\usepackage{setspace}
\usepackage{siunitx}
\usepackage[normalem]{ulem}

\graphicspath{ {./images/} }

\DeclareMathOperator*{\argmax}{arg\,max}

\newcommand{\trainingsizeperepoch}{10,000}
\newcommand{\trainingsizetotal}{one million}
\newcommand{\epochs}{100}
\newcommand{\testsize}{10,000}
\newcommand{\buffersize}{450}
\newcommand{\batchsize}{50}

\newcommand{\layers}{(256, 128, 64, 32)}
\newcommand{\dosingintervals}{the first two adjustments to be two and three days apart (on days three and six), and thereafter measure INR and adjust the dose every seven days}

\newcommand{\T}{90}
\newcommand{\ha}{1, 2, 3}

\newcommand{\discount}{0.95}
\newcommand{\learningrate}{0.001}

\newcommand{\explorationfn}{\ensuremath{(1+n)^{-1}}}

\newcommand{\mycolor}[1]{pink!#1}

\newcommand{\patientnormal}{\cellcolor{\mycolor{10}}}
\newcommand{\patientsensitive}{\cellcolor{\mycolor{50}}}
\newcommand{\patienthighlysensitive}{\cellcolor{\mycolor{100}}}

\begin{document}

\begin{frontmatter}

\title{Optimizing Warfarin Dosing using Deep Reinforcement Learning}

\author[1]{Sadjad Anzabi Zadeh\corref{cor1}}
\ead{sadjad-anzabizadeh@uiowa.edu}
\author[2]{W. Nick Street}
\author[3]{Barrett W. Thomas}
\address{Department of Business Analytics, Tippie College of Business, University of Iowa, Iowa City, IA 52242, USA}

\cortext[cor1]{Corresponding author}

\begin{abstract}
Warfarin is a widely used anticoagulant, and has a narrow therapeutic range. Dosing of warfarin should be individualized, since slight overdosing or underdosing can have catastrophic or even fatal consequences. Despite much research on warfarin dosing, current dosing protocols do not live up to expectations, especially for patients sensitive to warfarin. We propose a deep reinforcement learning-based dosing model for warfarin. To overcome the issue of relatively small sample sizes in dosing trials, we use a Pharmacokinetic/ Pharmacodynamic (PK/PD) model of warfarin to simulate dose-responses of virtual patients. Applying the proposed algorithm on virtual test patients shows that this model outperforms a set of clinically accepted dosing protocols by a wide margin. We tested the robustness of our dosing protocol on a second PK/PD model and showed that its performance is comparable to the set of baseline protocols.
\end{abstract}


\end{frontmatter}


\section{Introduction} \label{section: Introduction}

Accurate dosing of a drug is a crucial part of any treatment involving medications, particularly for chronic conditions that require continuous maintenance. The goal is to determine and adjust the dose of a drug to keep the patient in the therapeutic range and avoid Drug Related Problems (DRP), including over-dosages and therapeutic failures~\cite{krahenbuhl2007drug}. Homogeneity among humans justifies the use of population-based dosage calculations. However, inter-subject variability due to differences in factors such as genetic make-up of individuals, diet, and age results in difference in responses~\cite{maheshwari2018337}. This is more critical for Narrow Therapeutic Index (NTI) drugs, such as warfarin, and on complex patient populations~\cite{chan2017time}. The association between DRPs and NTI drugs are significantly higher than with non-NTI drugs, and non-optimal dosage is one of the top three concerns for NTI drugs~\cite{blix2010drugs}. In case of warfarin, DRP could be fatal, for an overdose increases chance of bleeding and an under-dose can result in thromboembolic events, in which clots formed in a blood vessel are carried by the blood stream and plug another vessel~\cite{shaw2015clinical}.

Building dosing protocols is a data-driven task that requires extensive experimentation and data gathering, both in model building and model validation, which makes extracting as much information as possible from clinical data in dosing studies a necessity. In common practice, dosing protocols are derived directly from empirical data. The study cohorts tend to be small due to time, cost and ethical considerations in medical research. Small cohorts adversely affect the model's performance at the population level. The other approach is to build mathematical dose-response models from observed data. This way, we enrich the data collected from a limited trial with the knowledge and insight from pharmacology and medicine, and extract a better dosing protocol. Moreover, these models can act as simulators that allow virtually an infinite number of observations and unobserved scenarios. Models resulting from such \emph{in silico} experiments are more generalizable to a broader population and are easier to validate. Moore~et~al.~demonstrates that a simulation-based dosing protocol can perform well on real patients, too~\cite{moore2014reinforcement}.

Building on successful examples of the application of Reinforcement Learning (RL) in drug dosing, like Moore~et~al.~\cite{moore2014reinforcement}, we find an optimal drug dosing protocol for warfarin. We chose a Pharmacokinetic/ Pharmacodynamic (PK/PD) model proposed by Hamberg~et~al.~\cite{hamberg2007pk} to simulate dose-responses. In a PK/PD model, the PK component models absorption, distribution, metabolism and elimination of the drug, and the PD component shows how the drug affects the organism~\cite{Alsanosi2014pharmacokinetic}. We model the optimization problem as a Markov Decision Process (MDP), and find an approximate solution to the MDP using RL. We also test the robustness of our dosing protocol by applying it to patients simulated using a second PK/PD model.

The rest of the paper is as follows: after a review of the relevant literature in warfarin dosing and sequential decision making in drug dosing in Section~\ref{section: Background and significance}, we present the problem description and its MDP in Section~\ref{section: Material and Methods}. Details of experiments and their results are provided in Sections \ref{section: Experiments}~and~\ref{section: Results}. We discuss some of the results in greater detail in Section~\ref{section: Discussion}, and conclude the study in Section~\ref{section: Conclusion}.

\section{Background and significance} \label{section: Background and significance}
More than six decades after its introduction, warfarin is still a widely-used anticoagulant. More than 8.7~million doses of warfarin were prescribed in the month of August in 2019 in the UK alone~\cite{ho2020trends}.While recently the number of warfarin prescriptions declined as a result of the introduction of a number of Direct-acting Oral Anticoagulants (DOACs), warfarin is the recommended choice or the preferred first-line treatment~\cite{ho2020trends}. The goal in warfarin dosing is to maintain blood coagulability of the patient, measured by a dimensionless factor called International Normalized Ratio (INR), in the therapeutic range---usually two to three. An INR above four increases the risk of bleeding and an INR lower than two increases the risk of thromboembolic events. Reported major bleeding incidents associated with warfarin are as high as 16\% and fatal incidents are as high as 2.9\%~\cite{shaw2015clinical}. Because of the narrow therapeutic range, high between- and within-patient variability, and serious side effects, warfarin dosing is challenging~\cite{pirmohamed2006warfarin}. As a result, some consider the problem of warfarin dosing as the most promising example of personalized medicine~\cite{fusaro2013systems}. 

We can loosely categorize warfarin dosing protocols into three categories. A majority of protocols, such as Gage et~al.~\cite{gage2008use} and Warfarin Pharmacogenetic Consortium (IWPC)~\cite{international2009estimation}, specify a maintenance dose of warfarin for each patient, a constant dose that a patient should take unless it requires adjustment. Most of these protocols are the result of statistical/supervised learning methods, especially regression models. While the predictors vary based on the collected data from the trial, the response variable is commonly the stable therapeutic dose, which is the dose that keeps the patient in the therapeutic range for a set number of days. The second category of protocols rely on adjustments in the maintenance dose. These are mainly simple rule-based models that are depicted in tabular format. INR values are divided into bins in these tables and the amount of dose change for each INR range is specified. 
An example is the IHC Chronic Anticoagulation Clinic Protocol Algorithm~\cite{anderson2007randomized}. Finally, there are algorithms to determine the initial dose(s) of warfarin, such as the pharmacogenetics-based 3-day warfarin initiation dose proposed by Avery~et~al.~\cite{avery2011a}. A combination of initial, adjustment, and maintenance protocols allows clinicians to put patients' INR in the therapeutic range and maintain their condition.

Protocols can also be differentiated based on the factors that they take into account. \emph{Clinical} dosing protocols include patient-specific factors, such as age, body surface area, and use of other medications to personalize dosing, but do not incorporate genetic information. However, genetic factors change patient's response to warfarin dramatically. In \emph{Pharmacogenetic} protocols, the genotypes or alleles that affect a patient's response are included. Among the genes that are identified as having an effect on warfarin sensitivity (reviewed systematically in~\cite{johnson2017clinical}), CYP2C9 (notably variants CYP2C9*2 and CYP2C9*3) and VKORC1 are the two most important. Shaw~et~al.~recommends testing of VKORC1, CYP2C9*2 and CYP2C9*3 for all patients~\cite{shaw2015clinical}. There are four main warfarin dosing algorithms, all based on regression, that incorporate genetic information: Gage~et~al.~\cite{gage2008use}, IWPC~\cite{international2009estimation}, Lenzini~et~al.~\cite{lenzini2010integration}, and EU-PACT~\cite{pirmohamed2013randomized}. The variables that each of these algorithms use are summarized in~\ref{appx: main_protocols_variables}.

The common use of regression models to build dosing protocols has a number of challenges and shortcomings. First, the cohort of patients in the trials for data collection is relatively small. For example, in one of the biggest attempts of its kind, IWPC collected data on more than 6,000 patients from 21 contributing sites. After data cleaning and removing records with missing information, the training set contained 4,043 patients and 1,009 patients were in the test set~\cite{international2009estimation}. Second, ethical considerations limit the doses that can be prescribed for patients. Inflexibility in decision making makes the outcomes sub-optimal, and the algorithm based on such data is less likely to produce an optimal dosing regimen. Finally, the resulting algorithms usually cover one phase of the dosing, initial, maintenance, or adjustments.

The alternative is to look at the problem of drug dosing as an instance of sequential decision making problems. We solve these problems using RL. Yu~et~al.~\cite{yu2019reinforcement} provide a review of RL application in healthcare. While RL applications in healthcare can be limited by data sufficiency and actions limitations due to ethical and practical constraints (see for example \cite{shortreed2011informing, levy2019applications, nemati2016optimal}), we avoid these limitation by performing \textit{in silico} experiments that allow RL methods to explore a virtually infinite number of patients and unobserved doses. This freedom cascades to the validation phase too, and we can validate models' performance unconditionally. An example of this approach to drug dosing is Padmanabhan~et~al.~(2017). They considered a nonlinear pharmacological model of cancer to develop a model-free RL model and validated its effectiveness~\cite{padmanabhan2017reinforcement}. Similarly, Humphrey used an ordinary differential equation model of cancer, and showed how different Q-function approximations can be employed to find the optimal dosing~\cite{humphrey2017using}. Other examples are reviewed in~\cite{yu2019reinforcement}.

Of course the resulting model will be as good in reality as the mathematical model can capture the reality of patients' response to medication. As mentioned in Section~\ref{section: Introduction}, examples such as Moore~et~al.~show that models developed in silico can perform well in actual trials as well~\cite{moore2014reinforcement}. A category of mathematical models that seem promising is PK/PD models. These models combine mathematical modeling techniques with the science of pharmacology to build models that explain how a medication is absorbed, distributed, metabolized and excreted in the body, and how the body responds to the medication~\cite{Alsanosi2014pharmacokinetic}. A PK/PD for warfarin allows us to experiment with different sequence of doses for different patients and see how they respond. The model that we picked is proposed by Hamberg~et~al.~\cite{hamberg2007pk}, and incorporates two genes (CYP2C9 and VKORC1) and the age of the patient. It captures randomness in response to warfarin between patients and between measurements. We also use a second PK/PD model, proposed by Hamberg et al.~\cite{hamberg2010pharmacometric}, to see how robust our proposed dosing protocol is when the underlying dynamics of dose-response changes.

To the best of our knowledge, RL has not been used in warfarin dosing. Although heparin as an anticoagulant is studied in Nemati~et~al.~\cite{nemati2016optimal}, it is not comparable to warfarin as they are administered differently and in different conditions. Moreover, the mentioned paper uses clinical data as we rely on a PK/PD model to generate data. We believe that the use of PK/PD not only eases the process of training and validating the model, but also leads to a more robust and effective dosing protocol since it adds our understanding of human body and metabolism to the observed data and produces a better sketch of the reality.

\section{Material and Methods} \label{section: Material and Methods}
In this section, we describe the process of warfarin dosing for a given patient. Then the problem formulation is presented. Finally, we discuss the solution method in detail.

\subsection{Problem Description} \label{subsection: Problem Description} 
In warfarin dosing, we need to define the patient's characteristics, possible decisions, decision points, and a performance measure to assess the effectiveness of a dosing protocol.

Our choice of the PK/PD model includes patient's age, CYP2C9, and VKORC1~\cite{hamberg2007pk}. Six genotypes of CYP2C9 and three genotypes of VKORC1 are included in the model. While in an in silico experiment, the distribution of patient characteristics can be left unconstrained, we wanted our cohort of virtual patients to better resemble a cohort of real warfarin recipients. To that end, we employed the characteristic information provided by Ravvaz et al. (2017). Specifically, Table 3 of that reference shows the extracted data from Aurora Health Care database of 14,206 patients with atrial fibrillation~\cite{ravvaz2017personalized}. It includes age, CYP2C9, and VKORC1 that are necessary for the PK/PD model, along with other characteristics, such as weight and height, that are needed for the baseline protocols. Table~\ref{tab: patient_parameters} shows all the adopted information. It is worth noting that we use these summary statistics to generate virtual patients, and since no living individual is involved about whom information is collected, used, studied, or analyzed, according to chart 1 of OHRP's ``Human Subject Regulations Decision Charts: 2018 Requirements'', this work is exempt from 45 CFR part 46 and does not require IRB approval. The choice of genotypes determine the distribution of parameters that describe the PK/PD model, such as volume of distribution, mean transit time, and apparent clearance (detailed description in~\cite{hamberg2007pk}). We do not include these in our state definition, as they are not observable. In other words, two virtual patients with the same age and genotypes will be different in their response to warfarin. We assume that none of the patient characteristics change for any given patient during the dosing trial.

The three characteristics are not sufficient to determine the dose properly, as they only define a patient and not the current state of the patient's health. We need to know the most recent observed INR. Moreover, we need to know the current concentration of warfarin in the patient's body and how the patient has responded to warfarin so far. Since these cannot be measured directly, we use a history of dosing decisions and INR measurements as indirect information. In many common protocols, such as Intermountain dosing protocol~\cite{anderson2007randomized}, the most recent dose and duration are implicitly included, as they adjust the dose as a percent of change in the dose, which requires knowing the previous dose. In our modeling, however, we treat the lengths of the decision and INR histories as a hyper-parameter to see if more historical information can have an impact on the performance of our model.

\begin{table}[!ht]
    \caption{Characteristics of the virtual patients}
    \label{tab: patient_parameters}
    \centering
    \begin{threeparttable}
        \begin{tabular}{l @{}S[table-format=3.2, table-column-width = 3.5cm]@{}}
            \toprule
            Characteristic  & {mean$\pm$SD} \\
            \midrule
            $Age$ (yr)\tnote{1} & 67.3$\pm$14.43 \\
            Weight (lb)\tnote{2} & 199.24$\pm$54.71 \\
            Height (in)\tnote{3} & 66.78$\pm$4.31 \\
            \rowcolor{lightgray} Sex (\%) & \\
            ~~~Female   & 53.14 \\
            ~~~Male     & 46.86 \\
            \rowcolor{lightgray} Race (\%) & \\
            ~~~White     & 95.18 \\
            ~~~Black     &  ~4.25 \\
            ~~~Asian     &  ~0.39 \\
            ~~~American Indian/ Alaskan     &  0.18 \\
            ~~~Pacific Islander     &  0.0001 \\
            \rowcolor{lightgray} Tobacco (\%) & \\
            ~~~No      & 90.33 \\
            ~~~Yes     & 9.66  \\
            \rowcolor{lightgray} Amiodarone (\%) & \\
            ~~~No      & 88.45 \\
            ~~~Yes     & 11.54  \\
            \rowcolor{lightgray} Fluvastatin (\%) & \\
            ~~~No      & 99.97 \\
            ~~~Yes     & 0.03  \\
            \rowcolor{lightgray} $CYP2C9$ (\%) & \\
            ~~~*1/*1     & 67.39 \\
            ~~~*1/*2     & 14.86 \\
            ~~~*1/*3     &  9.25 \\
            ~~~*2/*2     &  6.51 \\
            ~~~*2/*3     &  1.97 \\
            ~~~*3/*3\tnote{4}     &  0.00 \\
            \rowcolor{lightgray} $VKORC1$  & \\
            ~~~G/G       & 38.37 \\
            ~~~G/A       & 44.18 \\
            ~~~A/A       & 17.45 \\
            \bottomrule
        \end{tabular}
        \begin{tablenotes}
            \item[1] {Age is clipped to the range of [18, 100] based on a dataset of 10,000 virtual patients provided by~\cite{ravvaz2017personalized}.}
            \item[2] {Weight is clipped to the range of [70, 500] based on a dataset of 10,000 virtual patients provided by~\cite{ravvaz2017personalized}.}
            \item[3] {Height is clipped to the range of [45, 85] based on a dataset of 10,000 virtual patients provided by~\cite{ravvaz2017personalized}.}
            \item[4] In the implementation, we assumed the probability of observing this genotype to be $2.0\times 10^{-4}$.
        \end{tablenotes}
    \end{threeparttable}
\end{table}

Defining a decision point by superscript $n$, a dosing decision $x^n$ has two components: the dose $d^n$, and the duration $\tau^n$. Dose and duration act together to balance the frequency and effectiveness of interventions. The dose of warfarin should be discretized based on the standard tablet strengths. The tablets are color-coded and are available in 1, 2, 2.5, 3, 4, 5, 6, 7.5, and 10~mg. One can also split them in half if necessary~\cite{UCSD-tablet}. We use the range and values used by Fusaro~et~al. for warfarin doses~\cite{fusaro2013systems}. They assumed the maximum dose of 15.0~mg/day (although in rare cases the daily dose can be as high as 20.0~mg/day~\cite{johnson2017clinical}). The permutation of the standard doses gives us all values between zero and 15.0~mg with step of 0.5~mg. Most dosing protocols do not confine their choice of dose to standard tablet strengths. It is, therefore, the clinician's responsibility to specify each day's dose that is feasible and closest to the recommended dose~\cite{johnson2017clinical}. Note that implementing the clinician's role to get the dose values is necessary only for the baseline models, and require solving a small Mixed Integer Programming (MIP) model at each decision point. Hence, it is not a crucial component in our model, and we let baselines prescribe any real number for the dose.

The duration can be any number of days. We initially need to monitor the patient's INR more frequently and adjust the dose accordingly, so the prescribed duration of administering the drug is shorter. Having achieved a relatively stable dose-response, we can test and adjust the dose with less frequency~\cite{hirsh2003american}.
Despite the availability of point-of-care devices to measure INR by the patient since the 1990s, not many patients perform self-monitoring or self-management (15-20\% of patients in Netherlands, for example~\cite{biedermann2017optimization}), and fewer tests reduces the costs associated with care and improves patient experience. To avoid dealing with a large decision-space, we assume that the duration is determined beforehand as a sequence $(\tau^n)_{n\in\{1, 2,...\}}$ (discussed in Section~\ref{section: Experiments}). Day one is the first decision point, when the initial INR is measured. Then, we transition to next decision points according to the defined duration until the last day of the dosing trial $T$.

The transition from one state to the next requires measurement of INR. In our work, the PK/PD model computes the INR. Our choice of PK/PD is a two-compartment model with first-order input and first-order elimination, derived from data of 150 patients. It uses $E_{max}$ model for the response, and models the delay from medication intake to response using two parallel transit compartment chains~\cite{hamberg2007pk}.

For each patient, on day one, an instance of the PK/PD model is initialized by randomly generating the necessary parameters. Some of these parameters depend on patient characteristics, such as apparent oral clearance, and some are independent, such as mean transit times. At each decision point, the PK component calculates the change in warfarin concentration based on previous doses and the current dose. The PD component uses the computed concentration and calculates the INR for any queried day. Random noise is added to both concentration values and INR values to account for factors that are not part of the PK/PD model.

To better understand the dosing process, consider a 50-year-old patient, 5'4" height~($\approx$162~cm), weighing 182~lbs~($\approx$82.5~kg), with an initial INR of 1.3. Following the Aurora dosing protocol, described in~\cite{ravvaz2017personalized}, the physician prescribes 10~mg/day warfarin for days one and two. On day three, a new measurement shows INR increased to 1.8. Following the protocol, we need to increase the dose by 10\% and retest in seven days. The patient takes the new dose of warfarin, 11~mg/day, for seven days and retest shows his INR to be 3.1. Once again we should adjust the dose; this time decrease it by 10\% to 10.175 mg/day. The clinician would prescribe this new dose as 10.5~mg on Monday, Wednesday and Friday, and 10~mg for the rest of the week. After almost two months of adjustment, finally the patient is in the therapeutic range and we can do the test every month. If any monthly test shows an out-of-range INR, we need to adjust the dose again. This process continues for as long as warfarin is necessary for the patient.

If we use a pharmacogenetic dosing algorithm to initiate the trial, we can expect a better outcome. In our example, assume the two genotypes for the patient to be CYP2C9*1/*1 and VKORC1~G/A. The IWPC dosing protocol will recommend 4.99~mg/day for the first two days, which in practice will be one 5~mg tablet each day. Based on the new INR measurement of 1.6 and adjusting the dose according to Aurora protocol, the new dose will be 5.74~mg/day, and by the follow-up measurement in 7 days, the patient is in the therapeutic range.

The goal of dosing is to keep INR in a therapeutic range of two to three. We can measure this goal using Percent Time in Therapeutic Range (PTTR), which is the percent of days in the dosing period that the patient's INR was in therapeutic range. In the context of sequential decision making, \emph{reward} is the scalar value after each decision that signals how well the model is performing, and we want to maximize the cumulative reward. PTTR is not a good candidate for the reward function, because it cannot reflect trends in INR values. For example, constant in-range INR values have the same PTTR ($100\%$) as an upward trend from 2.0 to 3.0. Instead, we define a penalty for each day proportional to the Euclidean distance of that day's INR value from the mean of the therapeutic range. We normalize the penalty such that on the borders (INR values of 2.0 and 3.0), the penalty is $1.0$. As a result, the quadratic nature of this function puts more penalty on out-of-range INR values. The reward function is the negative of the total penalty between two decision points. Maximizing the reward will maximize PTTR as well.

In reality, only the INR values at measurement days are available to the decision maker. The unobserved values are equally important in a proper assessment of a patient's treatment, and computing the reward function in our model. Clinical trials estimate the unobserved INR values by simply interpolating the observed INR values, using Rosendaal method for example~\cite{rosendaal1993method}, which neglects the variations in daily INR. Even in silico trials tend to do the same, for example in~\cite{ravvaz2017personalized}. However, we can compute daily INR values using the PK/PD model, and acquire more information for solving our model. These INR values are not part of the state definition, because the decision maker should only be presented with observable information. Therefore, we treat daily INR values as exogenous information that are used solely to compute the reward function and measure the performance of different models. This is aligned with Powell's definition of a \emph{state variable}~\cite{powell2020reinforcement}.

\subsection{MDP Model} \label{subsection: MDP Model} 
In this section, we formulate the problem of warfarin dosing as an MDP. We define the decision points, the state space, the action space, the exogenous information, the transition function, the reward function, and finally the objective function.

\textit{Decision point}: The dosing trial starts with the first decision point $n=1$ on day $t=1$ with the initial INR measurement $\mu^1$. At each decision point $n$, the patient will take the medication with the prescribed dose $d^n$ for the duration $\tau^n$ prescribed by the model. At the next decision point, $n+1$, which is $\tau^n$ days after the $n^{th}$ decision point, a new INR, $\mu^{n+1}$, is measured. We make decisions until the time horizon $T$ is reached.

\textit{State} ($S$): The state is a tuple including the INR reading for the current decision point, a history of INR values and dosing decisions, and the static patient information. That is,
\begin{equation}
\label{eq: state}
S^n = \left(\mu^n, \mathcal{H}^n_h, P \right)
\end{equation}
\noindent
where $\mu^n$ is the latest INR value from the simulation of dose-response using the PK/PD, $P$ is the patient information defined as~$P=(Age, CYP2C9, VKORC1)$, and $\mathcal{H}^n_h$ is the list of the $h$ most recent observed INR values and their respective dosing decisions. The hyper-parameter $h$ determines how far in the past we should look as history. To have a fixed size history length, when $n<h$, we pad the lists with zeros for INR and dose, and ones for the duration.

\textit{Decision/Action} ($x$): The decision $x^n$ is defined as a tuple of $(d^n, \tau^n)$. Of the two components of the decision, the duration is pre-defined, and dose~$d^n$ is our decision variable. That is,
\begin{equation}
\label{eq: dose equation}
\begin{split}
    d^n&\in\mathcal{D}^n=\left\{0.0, 0.5, 1.0, 1.5, ..., D_{max}^n\right\}
\end{split}
\end{equation}
\noindent
where $D_{max}^n$ is the maximum allowed dose at decision point $n$. This value is 15~mg/day in our main model. However, as we will discuss in the experiments section, the maximum dose for the initiation of dosing might be lower due to safety concerns.

\textit{Exogenous information} ($W$): The exogenous information is the set of INR values we measure using the PK/PD model following the decision. It accounts for the change in the patient's INR, including the effect of drug concentration from previous doses, the effect of following the new prescribed dose for the determined duration, and random fluctuations due to factors such as diet and stress. The PK/PD incorporates the random factors as error terms for between-patient and between-INR measurements. We denote these INR values as $W^{n+1}=\left\{ \mu_1^n, \mu_2^n, ..., \mu_\tau^n \right\}$, where the subscript represents days after the decision point $n$. Of all these INR values, only the last one, $\mu_\tau^n$ is observable, and the rest are used to compute the reward and performance metrics.

\textit{Transition Function}: In the transition from the current state to the next, the patient information remains intact, the new INR value is $\mu^{n+1}=\mu_\tau^n$, and the history is updated by discarding the oldest record and adding the previous INR value, $\mu^n$, and the previous decision, $x^n$, to the history.

\textit{Reward}: Reward is a scalar value that indicates how much we have gained or lost following the dose~$d^n$ for the duration~$\tau^n$ in state~$S^n$ that transitioned us to the next state. As discussed in Section~\ref{subsection: Problem Description}, we use daily INR values from the PK/PD model to compute Euclidean distance as penalty, and the negation of the total penalty is our reward function. That is,
\begin{equation}
\label{eq: reward function}
    r(S^n, x^n,W^{n+1}) =-c \mathbb{E}\left[\sum_{t=1}^{\tau^n}{\left (\mu_m- \mu^n_t\right) ^ 2}\right]
\end{equation}
\noindent
where $\mu_m$  is the midpoint of the therapeutic range ($2.5$ in our case). Parameter $c$ is the normalization factor, and for our therapeutic range of two to three, $c=4$ normalizes the reward so that the reward for both INR values of two and three is $-1.0$. Note that in this formulation, we need the exogenous information revealed after the decision is made to compute the reward. For this reason, we have $W^{n+1}$ rather than $W^n$.

\textit{Objective Function}: Our objective is to maximize the total reward for all patients throughout the experiment period. We have:
\begin{equation}
\label{eq: simple obj}
    F^\star=\max_{\pi \in \Pi} \mathbb{E}\Big[\sum_{P\in \Phi}{\sum_{n}{r^\pi(S_P^n, x^{\pi,n}, W_P^{n+1})}\Big|S_P^0}\Big]
\end{equation}
\noindent
where $\Pi$ is the set of all possible policies, and $\Phi$ is the set of all patients.

\subsection{Deep Q-Learning} \label{subsection: Deep Q-Learning}
In our formulation, for each patient at each decision point, we need to choose a value from possible doses of warfarin. While the set of possible doses is fairly small and the combinations of genotypes produces a finite set of patients, the latent parameters that define a patient are random. Moreover, warfarin concentration and the resulting INR value after each dose have random fluctuations. As a result of these stochasticities, we cannot solve the model analytically. To approximate the optimal policy/ protocol, we use Q-learning. Q-Learning is a widely-used RL algorithm that learns the value associated with any pair of state and action, known as $Q$~value, in an iterative process that involves updating estimates of $Q$ values using Equation \ref{eq: q learning}, which is,
\begin{equation}
\label{eq: q learning}
    Q(S_i, A_i) \leftarrow Q(S_i, A_i) + \alpha (R_i + \max_a{Q(S_{i+1}, a)} - Q(S_i, A_i))\,.
\end{equation}
In this process, we take an action $A_i$ in a state $S_i$ and observe the reward $R_i$. When choosing the action, we need a trade-off between exploiting our current best action, based on our estimate of $Q$ values, and exploring other actions to improve $Q$ estimates and find potentially better decision sequences. We use an $\epsilon$-greedy approach \cite{powell2020reinforcement}, in which we deviate from the best action by some probability and choose a random action instead. The $Q(S_i, A_i)$ value should represent the observed reward $R_i$ and the $Q$ value of the immediate future state-action pair $(S_{i+1}, A_{i+1})$. So, we update our current estimate of $Q(S_i, A_i)$ by moving towards the maximum estimated new $Q(S_i, A_i)$ value, with step size $\alpha$. Values of $Q$ can be stored in a lookup table if the state-action space is small. In large spaces, however, we need to approximate $Q$ values as a function.

Neural networks (with at least one hidden layer) are long known to be capable of accurately approximating nonlinear functions~\cite{hornik1991approximation}, and new advancements in theory and computation capabilities has renewed the interest in employing deeper and larger neural networks. One issue with the use of deep neural networks in the RL domain is the correlation between observations that can result in high variance in updates and divergence. Experience replay solves this problem; instead of online updating, we keep a buffer of observations and update the neural network using a sampled mini-batch from the buffer~\cite{mnih2015human}. The depth (number of layers) of the neural network, the number of neurons in each layer, and other parameters such as size of the buffer and mini-batches are hyper-parameters that need to be set. In the next session we discuss the choices for these and other necessary parameters.

More details on the Q-learning algorithm, $\epsilon$-greedy method, and deep Q-learning can be found in~\cite{sutton2018reinforcement} and~\cite{powell2020reinforcement}.

\section{Experiments} \label{section: Experiments}
We implemented the PK/PD model and the deep RL model in Python, with TensorFlow for the neural network implementation. The RL model consists of a neural network that learns $Q$ values. It is a four-layer fully-connected network with ReLu activation function in hidden layers and sigmoid activation function in the output layer. The state-action-reward observations are stored in a buffer of size \buffersize, and the neural network is updated using a mini-batch of size \batchsize. More details on the deep Q-learning model are provided in Table~\ref{tab: model_parameters}. Note that our model is a finite horizon model, and we did not discount the reward in Equations~\ref{eq: simple obj} and~\ref{eq: q learning}. In practice, discounting the reward allows for faster convergence of $Q$ values, and we introduce the discount factor $\lambda$ in Table~\ref{tab: model_parameters}.

\begin{table}[!ht]
    \caption{Experiment setup}
    \label{tab: model_parameters}
    \resizebox{\textwidth}{!}{
    \centering
    \begin{tabularx}{\textwidth}{l l X}
        \toprule
        Parameter & Value & Description \\
        \midrule
        epochs & \epochs & The number of training iterations \\
        training\_size & \trainingsizeperepoch & The number of simulated patients used in each training iteration \\
        test\_size & \testsize & The number of simulated patients used as test set \\
        $T$ & \T & The duration of dosing for each patient (in days) \\
        $h$ & \ha & Number of previous INR and dosing decisions in state representation \\
        Q\_value\_updating & backward & whether to update Q-values in the forward pass or backward pass \\
        $\gamma$ & \discount & Discount factor of the future rewards \\
        $\alpha$ & \learningrate & The learning rate of the neural network \\
        buffer\_size & \buffersize & Number of observations stored for experience replay \\
        batch\_size & \batchsize & Size of the mini-batch sampled from the buffer \\
        hidden\_layer\_sizes & \layers & Number of neurons in each layer of the ANN \\
        exploration\_function & \explorationfn & The probability of randomly choosing an action during the training in epoch $n$ \\
        \bottomrule
    \end{tabularx}}
\end{table}

As discussed in the Section~\ref{subsection: MDP Model}, the duration part of each dosing decision is predetermined. To mimic the pattern of more frequent dosing in the initial phase and less frequent dosing in the maintenance phase, we assumed \dosingintervals.

We should also consider the loading dose in our experiments. In drug dosing, if no adverse effect is expected from large doses of a medication, clinicians would start with a loading dose, which is a large dose of the drug, to raise the concentration of the drug as fast as possible and put the patient measures in the therapeutic range~\cite{Alsanosi2014pharmacokinetic}. In the case of warfarin, however, high levels of warfarin can cause bleeding and possibly death. Therefore, the loading dose is usually limited to reduce such risks. CPIC guidelines recommend to use a pharmacogenetic protocol if a clinician wants to administer a loading dose~\cite{johnson2017clinical}. In practice, many dosing algorithms, such as Aurora dosing algorithm~\cite{ravvaz2017personalized}, start the dosing with a fixed 5 or 10~mg/day depending on the age and special medical conditions of the patient. We considered two scenarios with respect to the loading dose: unconstrained loading dose (15~mg/day similar to the rest of the dosing trial) and maximum initial dose of $D^1_{max}$=5~mg/day.

Like any other machine learning practice, we train our model, select the best trained model in the validation phase, and test it on a test set. At each training epoch, we generate \trainingsizeperepoch\ patients according to age and genotype probabilities depicted in Table~\ref{tab: patient_parameters} with 90-day dosing periods, and train for \epochs\ epochs (\trainingsizetotal\ distinct patients in total for training). We generate a new set of training patients in each epoch which avoids overfitting. Each trained model at the end of each epoch is a candidate for the best model.

To select the \emph{best} model from all \epochs\ trained models, we use a validation set of \testsize\ pre-generated patients. In this process, we should be aware of the fact that the genotypes of CYP2C9 and VKORC1 affect the sensitivity of patients (Table~\ref{tab: sensitivity}). it is more likely for a model to have better average performance (mean PTTR) on the cohort of normal patients than that of sensitive or highly sensitive. We also expect to see more variation (standard deviation of PTTR) in performance among highly sensitive and sensitive patients compared to normal patients. To ensure the safety and fair treatment of all patients, after each training epoch, we compute the mean and standard deviation of PTTR of each cohort separately, and consider the PTTR of the worst performing cohort as our measure of performance. Hence, the best model is selected based on Equation~\ref{eq: performance}:
\begin{equation}
\label{eq: performance}
    \argmax_{m=1}^{\epochs}{\min_{s\in \mathcal{S}}{\overline{PTTR}^m_s - {SD}_{PTTR^m_s}}}
\end{equation}
\noindent
where $m$ denotes the trained model at each epoch, $s$ is the sensitivity level, and $\mathcal{S}$ is the set of all sensitivity levels.

The selected model is then used for the test phase, for which a set of \testsize\ pre-generated patients is used. We report the mean and standard deviation of PTTR values for the test set, segmented by sensitivity levels. All the training, validation, and test phases are done using the PK/PD introduced in Sectoin~\ref{section: Background and significance}. In the discussion section, we test our proposed dosing protocol on a separate PK/PD (Hamberg et al. 2010~\cite{hamberg2010pharmacometric}) to demonstrate the robustness of our approach.

\begin{table}[!ht]
    \caption{Patient sensitivity based on genotypes~\cite{ravvaz2017personalized}}
    \label{tab: sensitivity}
    \centering
    \begin{tabular}{c c c c c c c}
        \toprule
         VKORC1 & \multicolumn{6}{c}{CYP2C9} \\
         				  & *1/*1 & *1/*2 & *1/*3 & *2/*2 & *2/*3 & *3/*3 \\
         \midrule
         G/G & \patientnormal & \patientnormal & \patientsensitive & \patientsensitive & \patientsensitive & \patienthighlysensitive \\
         G/A & \patientnormal & \patientsensitive & \patientsensitive & \patientsensitive & \patienthighlysensitive & \patienthighlysensitive \\
         A/A & \patientsensitive & \patientsensitive & \patienthighlysensitive & \patienthighlysensitive & \patienthighlysensitive & \patienthighlysensitive \\
         \bottomrule
    \end{tabular}
\mbox{\begin{tabular}{llllll}
\textcolor{\mycolor{10}}{$\blacksquare$} & Normal &
\textcolor{\mycolor{50}}{$\blacksquare$} & Sensitive &
\textcolor{\mycolor{100}}{$\blacksquare$} & Highly Sensitive \\
\end{tabular}}
\end{table}

\begin{table}[!ht]
    \caption{Baseline dosing algorithms~\cite{ravvaz2017personalized}}
    \label{tab: baseline_dosing_algorithms}
    \centering
    \begin{tabular}{l l l l}
        \toprule
         Name                               & Initial (Day)         & Adjustment (Day)  & Maintenance (Day) \\
         \midrule
         Fixed-dose protocol AAA          & Aurora (1-2)          & Aurora (3-7)      & Aurora (8-90) \\
         Clinically guided protocol CAA   & IWPC Clinical (1-2)   & Aurora (3-7)      & Aurora (8-90) \\
         PG-Guided protocol PGAA          & IWPC PG (1-3)         & Aurora (3-7)      & Aurora (8-90) \\
         PG-Guided protocol PGAI          & Modified IWPC PG (1-3)& Lenzini PG (4-5)  & Intermountain (6-90) \\
         PG-Guided protocol PGPGA         & Modified IWPC PG (1-3)& Lenzini PG (4-5)  & Aurora (6-90) \\
         \bottomrule
    \end{tabular}
\end{table}

To compare our proposed model with current dosing protocols, we adopted the study arms of Ravvaz~et~al.~as our baselines~\cite{ravvaz2017personalized}. Details of each of these baselines are presented in Table~\ref{tab: baseline_dosing_algorithms}. Each baseline is a composition of three protocols for \emph{initial}, \emph{adjustment}, and \emph{maintenance} phases. The first protocol, AAA, employs ``Aurora best-practice standard dose warfarin therapy protocol''~\cite{ravvaz2017personalized} throughout the dosing period. CAA determines the initial dose using ``IWPC clinically-guided algorithm''~\cite{international2009estimation}, and the rest is based on the Aurora protocol. The third baseline, PGAA, incorporates the genetic information by using ``IWPC PG-guided algorithm''~\cite{international2009estimation} in the initial phase, and Aurora for the rest of the duration. The last two baselines initialize based on IWPC-PG, and then adjust the dose according to ``Lenzini PG-guided algorithm''~\cite{lenzini2010integration}. The difference between the two is in the maintenance phase, in which PGPGI uses ``INR-based Intermountain Healthcare Chronic Anticoagulation Clinic Protocol''~\cite{anderson2007randomized}, while PGPGA follows the Aurora protocol.  Ravvaz~et~al.~also computed a simple complexity score for each protocol and rank them accordingly. Based on this ranking, AAA is the simplest protocol followed by CAA, PGPGA, PGPGI, and finally PGPGA.

Along with PTTR as our main performance measure, we also compare models based on measures described in Table~\ref{tab: perf_measures}.

\begin{table}[!ht]
    \caption{Performance measures}
    \label{tab: perf_measures}
    \centering
    \begin{threeparttable}
        \begin{tabular}{p{0.35\linewidth}p{0.55\linewidth}}
            \toprule
             Name                              & Description \\
             \midrule
             Mean number of decision points  & Average number of decisions made per person during the dosing trial \\
             Mean time to first therapeutic     & Average number of days since the start of the dosing to get to therapeutic INR \\
             Mean daily dose to first therapeutic    & Average daily dose since the start of the dosing to get to therapeutic INR \\
             Mean daily dose post first therapeutic   & Average daily dose since the first therapeutic INR \\
             Mean daily dose total                    & Average daily dose during the dosing trial \\

             \bottomrule
        \end{tabular}

    \end{threeparttable}
\end{table}

For sensitivity analysis of our proposed method, we run different experiments. More specifically, we are interested to see (a) how change in the history length $h$ affects the performance of the model, and (b) if our model can have a comparable performance when genotype information is not available. Both of these experiments have practical significance as well. The history length analysis helps us understand if we can still use this algorithm for a patient with little or no prior dosing information. The genotype information availability is also an issue to consider. Oftentimes, genotypes of a patient are not determined in the first few days of dosing, and the intervention cannot be postponed until the information is available~\cite{lenzini2010integration}. We also should not assume that genotyping is available for all patients, especially in less affluent communities and countries.
\section{Results} \label{section: Results}
In this section, we compare the results of our proposed model with baseline protocols. Then we discuss the sensitivity analysis and how different factors affect our model's performance. The last part of the Results section shows how our proposed model is robust to the changes in dose-response dynamics.

\subsection{Proposed model vs. baseline protocols}
Table~\ref{tab: pttr_base_vs_baselines} shows the mean and standard deviation (in parentheses) of the PTTR of our base model ($h=1$ ``with genotype'') along with baselines' statistics. Our trained model outperforms the baselines for all sensitivity levels. The difference in all sensitivity levels is statistically significant (t-test with $95\%$ significance level). The gap is greater for sensitive and highly sensitive groups and for simpler baseline protocols. The trained model has a lower variation in each sensitivity group as well. The best PTTR in the baseline group belongs to PGAA protocol for normal patients with mean PTTR of $78.5\%$ and standard error of $0.13$. These numbers for the base model are $92.4\%$ and $0.04$, respectively. For the 90-day period of trials, our model's PTTR values indicate that normal, sensitive and highly sensitive patients are on average 83, 80, and 81 days in the therapeutic range.

\begin{table}[!ht]
    \caption{Percent Time in Therapeutic Range of our model vs. baseline protocols}
    \label{tab: pttr_base_vs_baselines}
    \centering
    \begin{tabular}{l c c c c c c}
        \toprule
        protocol    & Base model                 & AAA          & CAA            & PGAA          & PGPGA         & PGPGI \\
        sensitivity \\
        \midrule
        normal      & \textbf{92.4\% (0.04)} & 72.4\% (0.18) & 73.9\% (0.17) & 78.5\% (0.13) & 74.8\% (0.15) & 59.1\% (0.32) \\
        sensitive   & \textbf{89.3\% (0.11)} & 43.8\% (0.27) & 60.3\% (0.26) & 68.5\% (0.23) & 73.0\% (0.20) & 63.5\% (0.30) \\
        highly sens.& \textbf{90.5\% (0.08)} & 14.2\% (0.15) & 24.9\% (0.21) & 59.1\% (0.25) & 55.4\% (0.26) & 43.4\% (0.33) \\
        \midrule
        all         & \textbf{91.3\% (0.08)} & 60.3\% (0.27) & 67.3\% (0.23) & 74.3\% (0.18) & 73.4\% (0.18) & 60.0\% (0.32) \\
        \bottomrule
    \end{tabular}
\end{table}

Table~\ref{tab: doses_base_vs_baselines} presents the mean and standard deviation (in parentheses) of the average prescribed dose by our protocol and baseline protocols. In our protocol, the average dose for normal and sensitive patients are 3.8 times and 1.9 times the average dose for highly sensitive patients, which are the biggest differences in average doses due to sensitivity. The standard deviations are also higher than baselines. These values reflect the ability of the protocol to differentiate patients, as the trajectories of doses and INR values indicate as well.

\begin{table}[!ht]
    \caption{Mean and standard deviation of the average prescribed doses by each protocol}
    \label{tab: doses_base_vs_baselines}
    \centering
    \begin{tabular}{l c c c c c c}
        \toprule
        protocol    & Base model & AAA           & CAA           & PGAA          & PGPGA         & PGPGI \\
        sensitivity \\
        \midrule
        normal      & 8.36 (3.39)	& 8.38 (2.75)	& 6.53 (1.65)	& 7.19 (1.90)	& 5.61 (1.38)	& 6.16 (1.74) \\
        sensitive   & 4.21 (2.79)	& 5.93 (2.61)	& 4.88 (1.49)	& 4.27 (1.29)	& 3.75 (0.83)	& 3.85 (0.99) \\
        highly sens.& 2.18 (2.31)	& 4.04 (2.73)	& 3.39 (1.60)	& 2.33 (0.82)	& 2.55 (0.64)	& 2.43 (0.74) \\
        \bottomrule
    \end{tabular}
\end{table}

\begin{figure}[h]
    \centering
    \includegraphics[width=\textwidth]{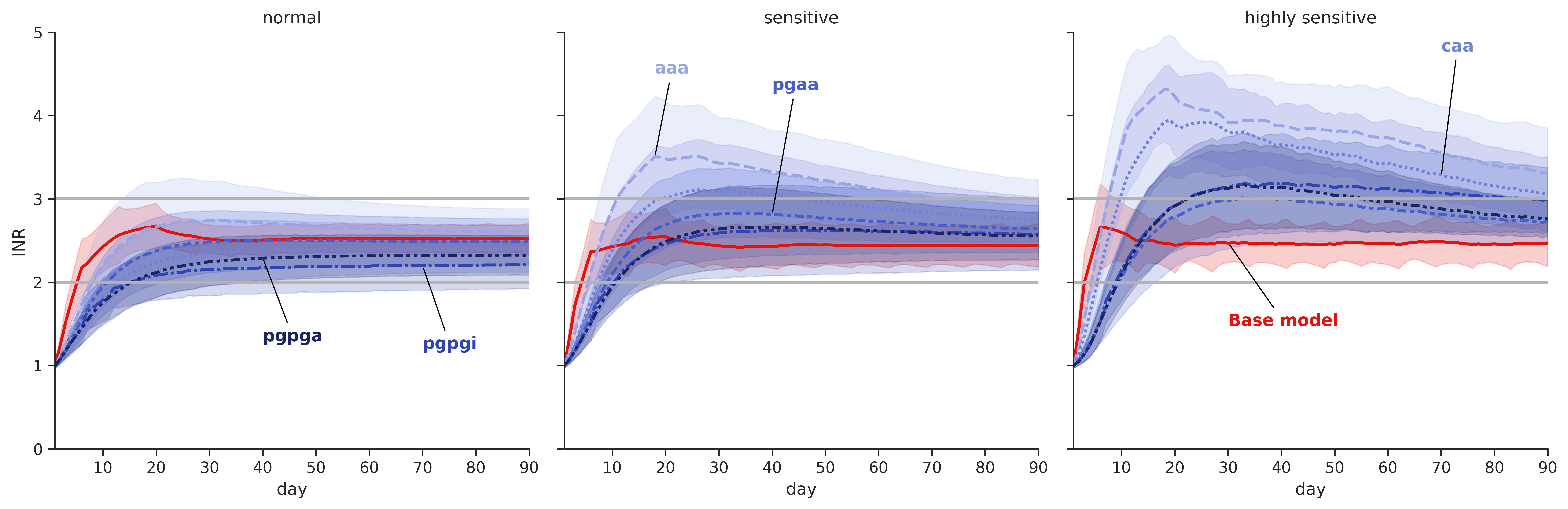}
    \caption{Mean and standard deviation (shaded areas) of INR values on each day}
    \label{fig: inr_comparison}
\end{figure}

Studying the trajectory of doses and INR values can help us better understand why the proposed method is highly effective. The INR values over the dosing trial time is depicted in Figure~\ref{fig: inr_comparison}. Sub-plots represent patients at different sensitivity levels, with INR values on the vertical axes and days of the trial on the horizontal axes. Each line shows the average INR value at each day following each protocol, and shades show the standard deviations. The therapeutic range of 2.0 to 3.0 is marked by gray horizontal lines. INR values for all patients at the start of the trial is close to one, which is the INR of a normal person. Then the values increase gradually according to the prescribed dose and the response of the patient. All baseline protocols do a similar job on normal patients and can transition and keep patients in the therapeutic range, yet with more delay and more chance of out-of-range INR values. The difference is more obvious when comparing performance on highly sensitive patients. All baseline protocols overshoot initially and then aim for the therapeutic range gradually. Our proposed model, however, has no problem aiming at the right target. 

\begin{figure}[h]
    \centering
    \includegraphics[width=\textwidth]{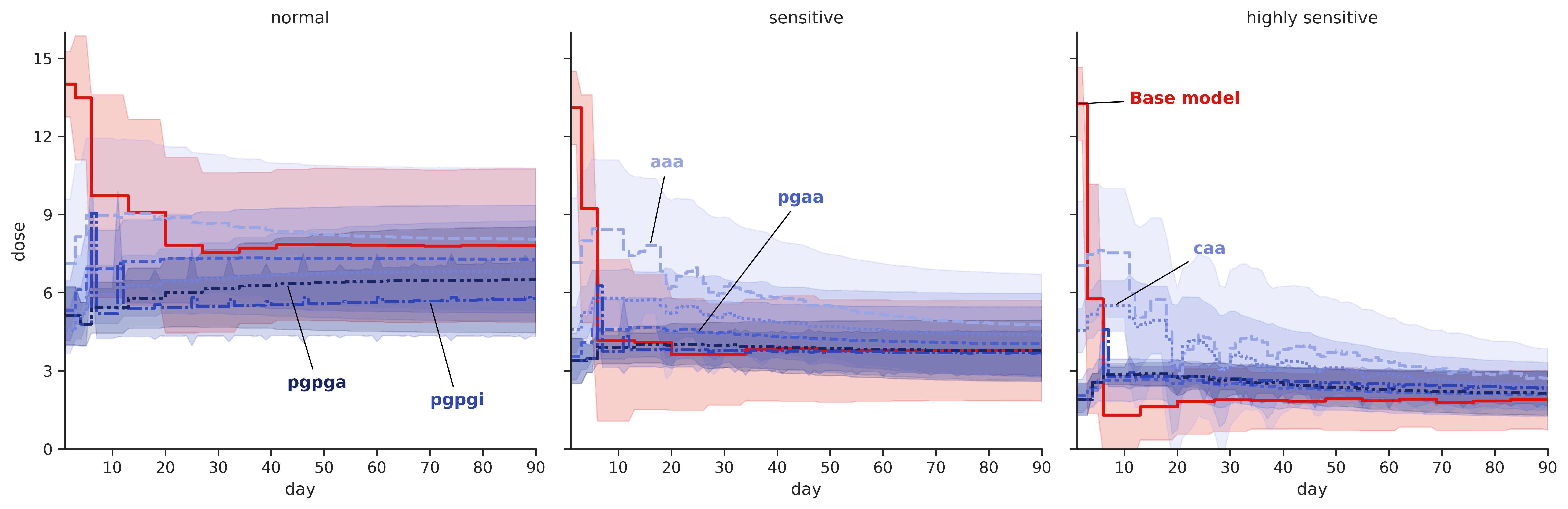}
    \caption{Mean and standard deviation (shaded areas) of dose values on each day}
    \label{fig: dose_comparison}
\end{figure}

Dosing trends (Figure~\ref{fig: dose_comparison}) reveal how the baseline protocols fail. In this figure, vertical axes are dose values and each line represents the average prescribed dose by each protocol. Breaks in each line corresponds to a dosing decision. For the baseline protocols, dosing days vary depending on the response of each individual. Initially, all baseline protocols underdose and delay patients' transition to the therapeutic range. Moreover, ``Aurora'' and ``Intermountain'' protocols, which are used in all baselines for the maintenance phase of the dosing, are not flexible enough to change the dose drastically. The gradual and days-apart changes to the dose fails to accelerate/ decelerate the change in INR in a timely fashion. In contrast, our proposed model has learned that patients initially need a high amount of warfarin to achieve the therapeutic range, and then a small dose can maintain the therapeutic effect.

\subsection{Sensitivity Analysis}
In the sensitivity analysis, we considered three values for the history length $h$ and two cases of genotype information availability. We also trained a model with reduced loading dose. Table~\ref{tab: pttr_base_vs_variations} show the mean and standard deviation (in parentheses) of PTTRs for each protocol. The case of ``with genotype'' and $h=1$ is the base model. As expected, limiting the loading dose (doses for days one and two) to 5~mg adversely affects the performance. The drop in PTTR values from the base model are minimal, but statistically significant. The $D_{max}^1=5mg$ model still manages to show PTTR numbers higher than all baselines.

Increasing history length $h$ is expected to improve the performance of the model. However, for $h=2$, the numbers show that it is the worst performing of our dosing protocols. Study of different aspects of the model did not reveal why this pattern exists. Genotype availability cases also present a surprising but consistent pattern: lack of genotype information improves the performance both in terms of the mean PTTR and the standard deviation, especially for sensitive and highly sensitive patients. We used one-tail and two-tail paired-t-tests ($\alpha=0.05$) to rank all the trained models. The ranking from the best performing to the worst performing models are as follows:

\begin{itemize}
    \item $h=3$ without genotypes
    \item $h=2$ without genotypes
    \item $h=3$
    \item $h=1$ without genotypes
    \item $h=1$ \\
          $D_{max}^1=5.0 mg$ without genotypes
    \item $D_{max}^1=5.0 mg$
    \item $h=2$
\end{itemize}

\begin{table}[!ht]
    \caption{Percent Time in Therapeutic Range in sensitivity analysis models}
    \label{tab: pttr_base_vs_variations}
    \centering
    \begin{tabular}{l l c c c c}
        \toprule
        Parameter      &                   & $h=1$         & $h=2$         & $h=3$         & $D_{max}^1=5mg$ \\
        genotype availability   & sensitivity \\
        \midrule
        with genotypes      & normal            & 92.4\% (0.04)	& 90.9\% (0.06)	& 92.5\% (0.04)	& 91.0\% (0.05) \\
                            & sensitive         & 89.3\% (0.11)	& 87.7\% (0.11)	& 92.6\% (0.05)	& 89.4\% (0.06) \\
                            & highly sensitive  & 90.5\% (0.08)	& 85.0\% (0.08)	& 92.7\% (0.05)	& 87.6\% (0.08) \\
                            & all               & 91.3\% (0.08)	& 89.6\% (0.08)	& 92.6\% (0.04)	& 90.3\% (0.06) \\
        \midrule[1pt]

        without genotypes   & normal            & 90.8\% (0.04)	& 92.7\% (0.03)	& 92.9\% (0.03)	& 91.1\% (0.04) \\
                            & sensitive         & 93.3\% (0.03)	& 93.2\% (0.04)	& 93.7\% (0.04)	& 91.5\% (0.05) \\
                            & highly sensitive  & 94.0\% (0.04)	& 92.6\% (0.06)	& 94.0\% (0.04)	& 92.0\% (0.04) \\
                            & all               & 91.8\% (0.04)	& 92.9\% (0.04)	& 93.2\% (0.03)	& 91.3\% (0.04) \\
        \bottomrule
    \end{tabular}
\end{table}

The difference between the best and the worst trained models is 3.6 percentage points. Other than the case of $h=2$, limiting the loading dose has the most significant effect, since it does not allow the initial boost to get the patient in the therapeutic range. All ``without genotype'' models perform better than their ``with genotype'' counterparts. This can be explained by studying two aspects of the training process: sampling and model selection. In this work, we use the same sampling distribution for training as we use for the test. Some of the genotypes that characterize sensitive and highly sensitive patients are quite rare or infrequent. As a result, the neural network does not see enough samples of these genotypes to learn how to deal with them. Specifically, the weights of the first layer that correspond to these genotypes are not fully learned in the training process. Therefore, when a patient is of any of these rare genotypes, the genotype information negatively impacts the decision making.

Second, we select the best model as explained in Section~\ref{section: Experiments}. In this approach, PTTR distribution of highly sensitive group often becomes the decisive factor since this group has higher variance and lower mean for PTTR. In $h=1$ for example, in 58 out of 100 training iterations, the mean and standard deviation of PTTR for normal patients in the validation set is $\ge90.0\%$ and $<0.10$. For sensitive and highly sensitive patients, this number is 21 and one iterations, respectively. The same cannot be said about ``no genotype'' models, since the models learn to treat all patients according to their response to warfarin, and they do not suffer from not-trained weights corresponding to genotype information. In $h=1$ ``without genotype'' case, for example, 78 iterations for normal, 73 iterations for sensitive, and 25 iterations for highly sensitive patients have PTTR of $\ge90\%$ with standard deviation of $<0.10$. One remedy to both of these issues is oversampling rare genotypes during the training process, which will be discussed in Section~\ref{section: Discussion} in more detail.

\begin{figure}[h!]
    \centering
    \includegraphics[width=0.94\textwidth]{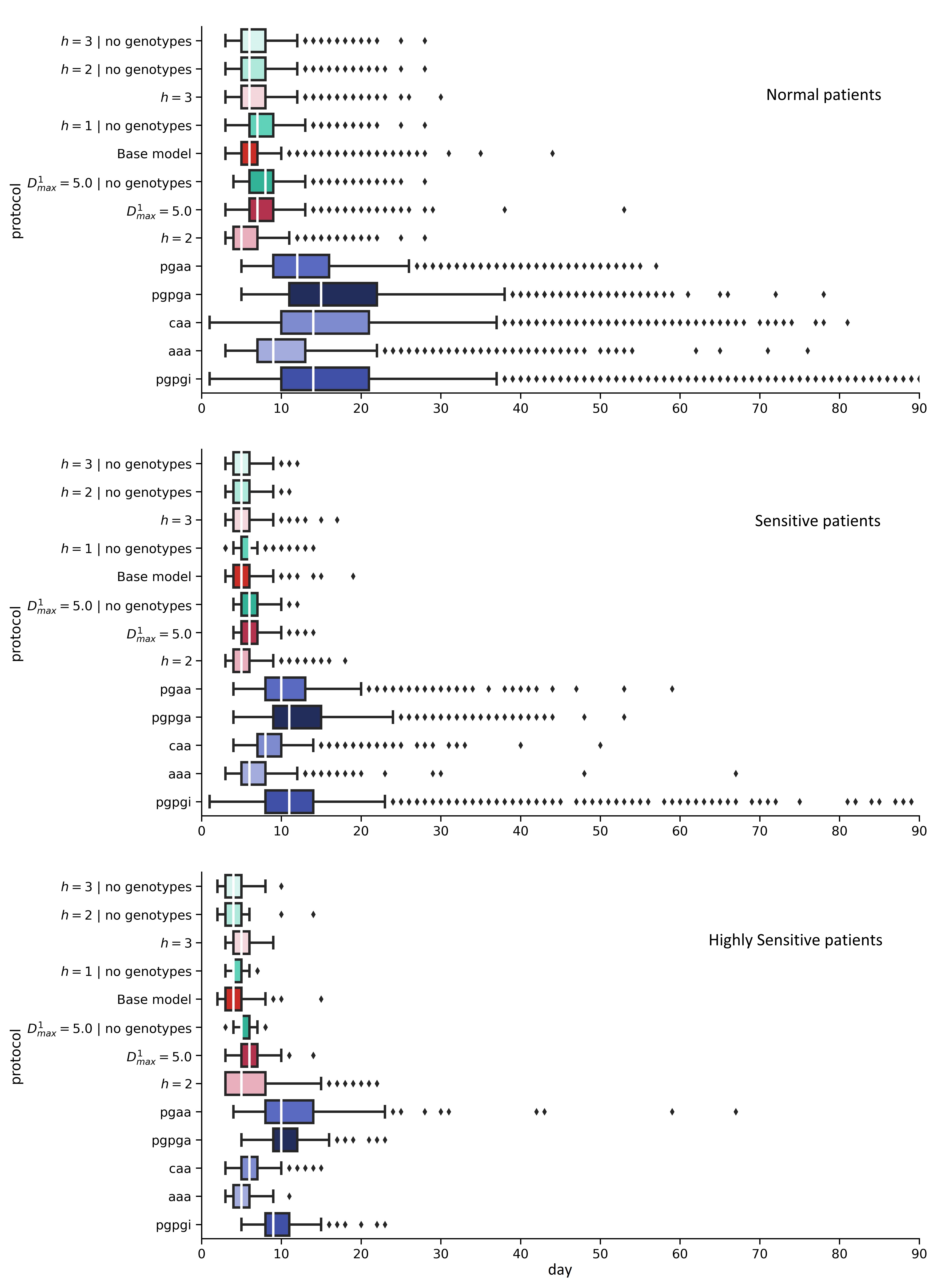}
    \caption{Number of days to first therapeutic-level INR}
    \label{fig: first_ttr_boxplots}
\end{figure}

\begin{figure}[h!]
    \centering
    \includegraphics[width=0.85\textwidth]{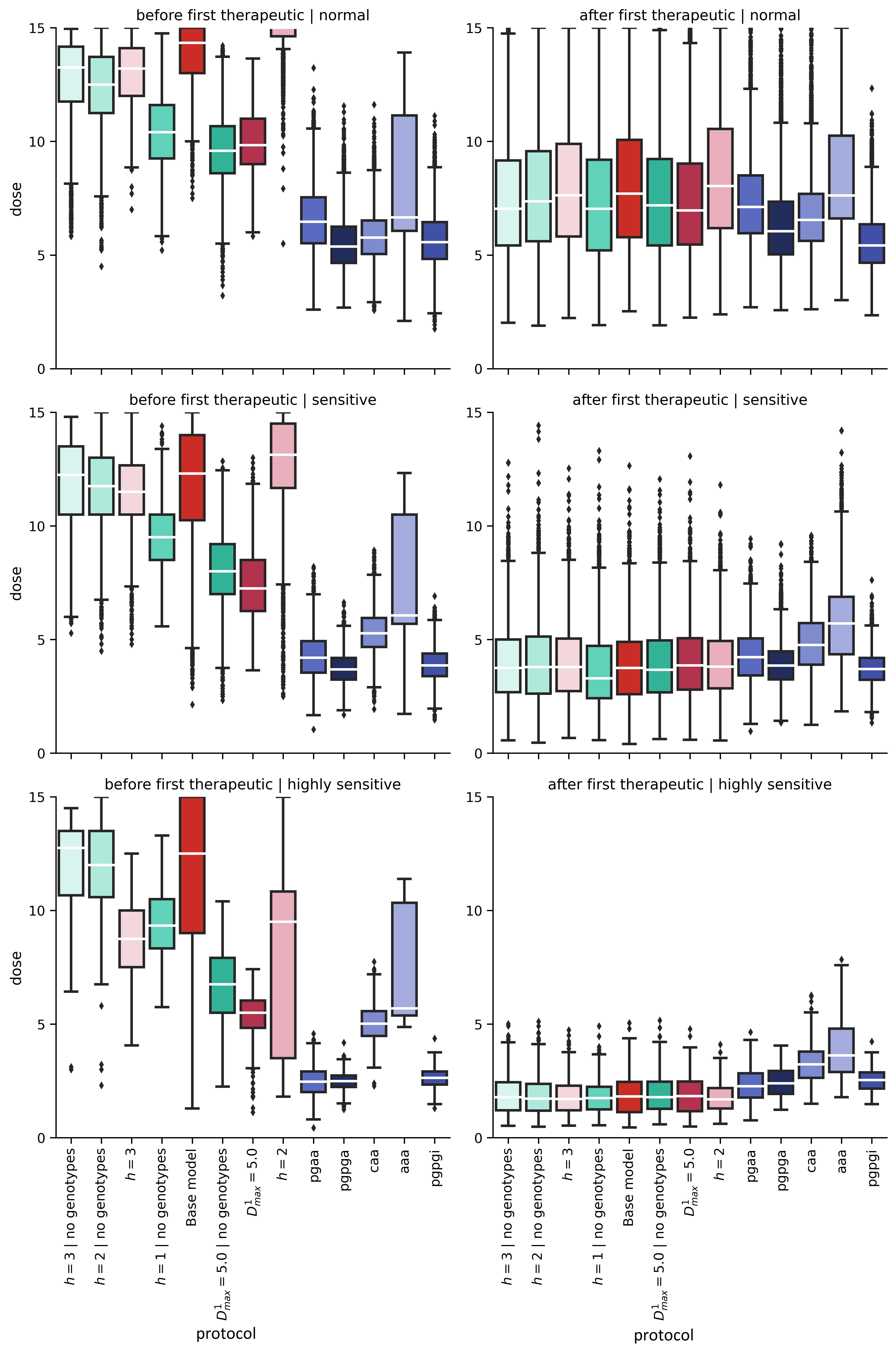}
    \caption{Average daily dose before and after the first therapeutic day}
    \label{fig: pre_post_dose_boxplots}
\end{figure}

Other performance measures are presented in Table~\ref{tab: decision_points} and Figures~\ref{fig: first_ttr_boxplots}~and~\ref{fig: pre_post_dose_boxplots}. As discussed in Section~\ref{subsection: Problem Description}, our proposed models have a fixed number of decision points. The baseline protocols adjust the number of decision points based on patients' response. Since INR measurement usually requires lab work, the baseline protocols are on average more comfortable for patients. Table~\ref{tab: decision_points} shows the average number of dosing decisions along with the standard deviations in parentheses.

\begin{table}[!ht]
    \caption{Mean and standard deviation of the average number of decision points by each protocol}
    \label{tab: decision_points}
    \centering
    \begin{tabular}{l c c c c c c}
        \toprule
        protocol    & Proposed models & AAA           & CAA           & PGAA          & PGPGA         & PGPGI \\
        \midrule
        decision points      & 15.0  & 14.10 (4.15)  & 12.72 (3.41)  & 11.83 (2.97)  & 11.61 (2.70)  & 9.98 (2.57) \\
        \bottomrule
    \end{tabular}
\end{table}

Figure~\ref{fig: first_ttr_boxplots} shows the distribution of the first day patients get into the therapeutic range under each dosing protocol. Protocols are shown in the order of their performance on the vertical axes, and horizontal axes are days. Each boxplot shows the inter-quartile range, with median day marked with a white vertical line, and whiskers extending to 1.5 the IQR. The outliers can represent more than one patient. In general, models with better PTTR values spend shorter time to get patients into the range. However, this is not always the case. For example, ``AAA'' is the fastest among the baseline protocols to get patients in the range, but is the worst baseline protocol in terms of PTTR measure.

Figure~\ref{fig: pre_post_dose_boxplots} shows how average daily dose changes prior to the first therapeutic day and after that. The vertical axes are daily doses and each box corresponds to a protocol. The plots on the left side depict boxplots of the doses before the first therapeutic day, with median as a white line inside the quartile ranges. The right-side plots show the average daily dose after patients entered the therapeutic range. While average daily dose after the first therapeutic day is relatively the same across our trained models, the dosing prior to that day is different among protocols. Our base model and all $h$ models, except for $h=2$ have higher pre-therapeutic doses, but their average daily dose after the first therapeutic day is comparable to baseline protocols. Among the trained models, ``no genotype'' variants have smaller IQR, which can be attributed to the fact that these protocols do not have access to genotype information, and are not able to differentiate between patients based on their genetic makeup. 

\begin{figure}[h]
     \centering
     \begin{subfigure}[b]{0.45\textwidth}
         \centering
         \includegraphics[width=\textwidth]{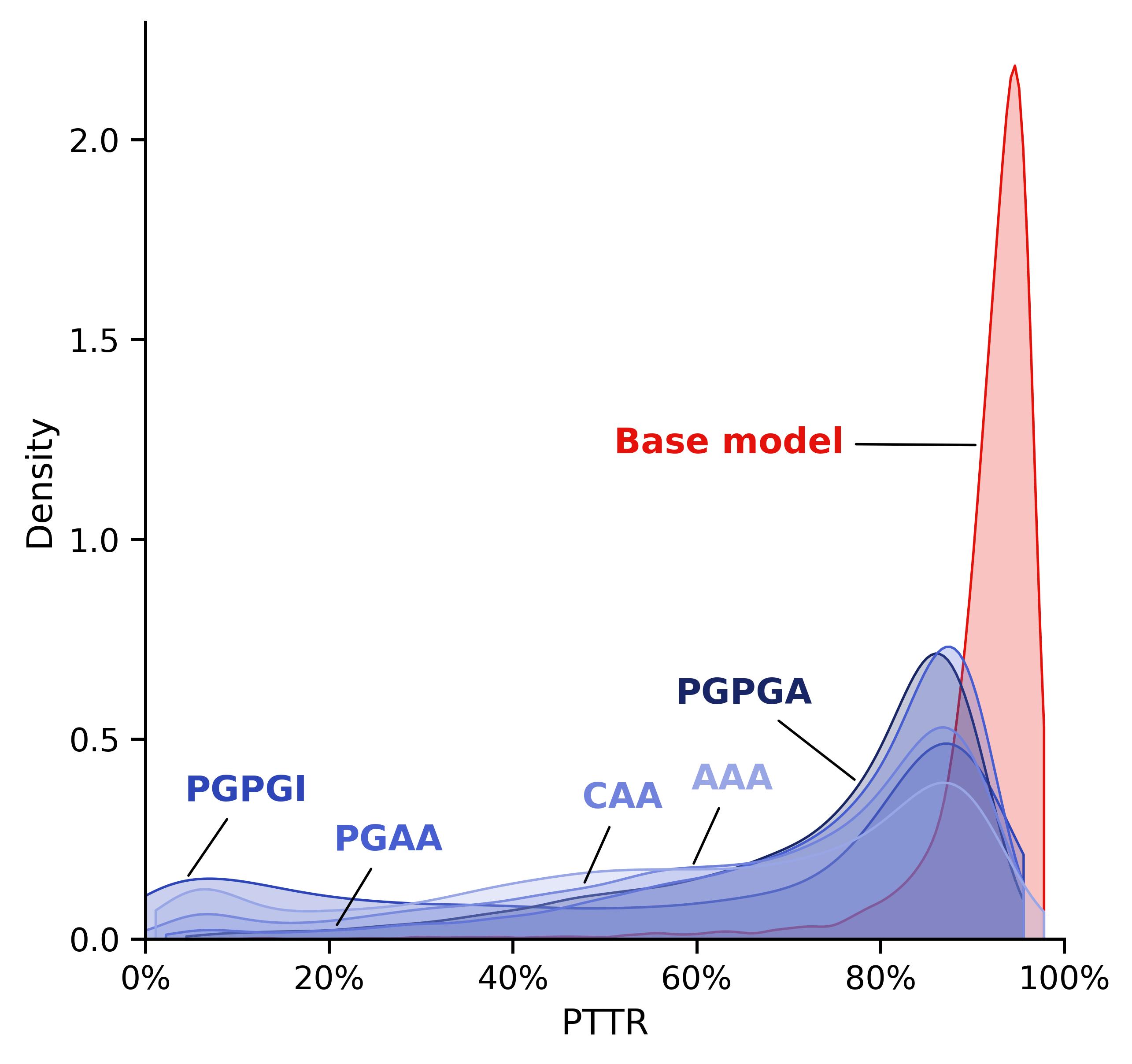}
         \caption{The distribution of PTTR values for the base model and baseline protocols}
         \label{fig: PTTR distributions}
     \end{subfigure}
     \hfill
     \begin{subfigure}[b]{0.45\textwidth}
         \centering
         \includegraphics[width=\textwidth]{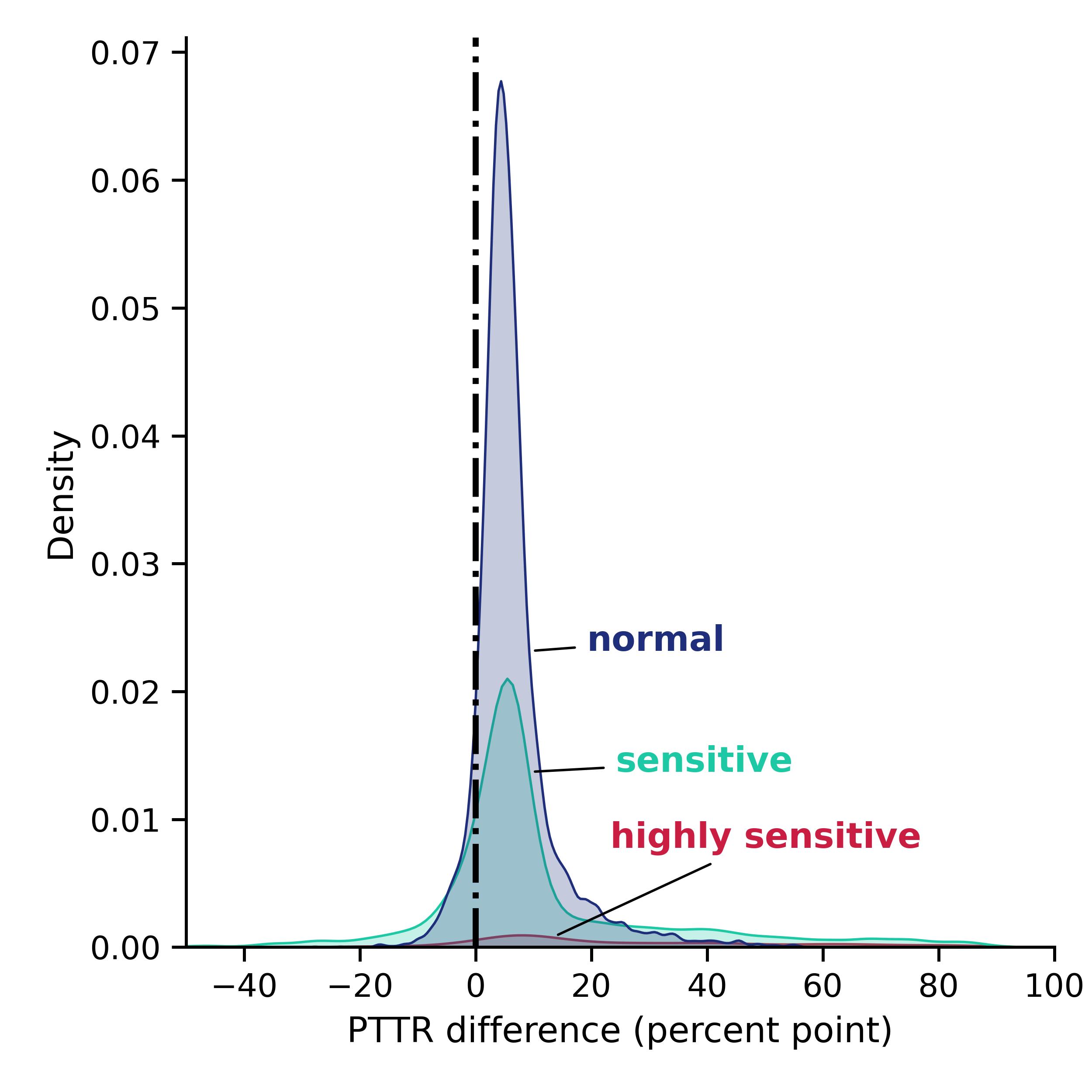}
         \caption{Percentage point gain/loss of individuals by adopting the proposed protocol vs their best respective baseline protocol}
         \label{fig: PTTR differences}
     \end{subfigure}
        \caption{Comparison of PTTR values}
        \medskip
        \label{fig: PTTR figures}
\end{figure}

The next question to discuss is whether the proposed method helps all individuals. We saw that PTTR values of our work is much better than the baselines, but it might not be the case for all patients at individual level. Figure~\ref{fig: PTTR distributions}) shows the distribution of PTTR values for the test patients following each dosing protocol. The horizontal axis shows PTTR values and the vertical axis shows the density. The distribution of PTTRs for our base model, shown in red, confirms that most of the test patients can benefit from our proposed model. But we need to consider each and every patient if we want to fairly personalize the treatment. If we take the best baseline protocol for each individual and compare its corresponding PTTR value with that of our base model, we can see the gain or loss in PTTR at individual level. Figure~\ref{fig: PTTR differences} shows the distribution of this change in PTTR. The horizontal axis show the amount of change in PTTR if the patient follows our protocol compared to the best baseline for that patient. The dashed vertical line marks the point zero, which indicates equal performance. For the majority of patients in all three sensitivity levels, we see improvement in their time in therapeutic (the area to the right of the dashed vertical line). However, there are cases for which our proposed method is not as effective as the best baseline protocol. We see that 6.9\%, 16.5\%, and 9.1\% of normal, sensitive, and highly sensitive patients (1,033 patients out of 10,000 test patients) can achieve higher PTTR values in one or more of baseline protocols compared to our base model. The median change in PTTRs are 3.3, 5.6, and 6.7 percentage points for each sensitivity group, respectively (equivalent of 3, 5, and 6 more days in therapeutic range). If we consider our best model, $h=3$ ``without genotypes'', only 4.2\% of normal and 2.7\% of sensitive patients do not benefit from our work (351 patients total). The median change in PTTR in this case for both sensitivity levels is 2 days. All highly sensitive patients in this case benefit from our dosing protocol. We study few notable examples of patients on both ends of the distribution in~\ref{appx: individuals}. We also studied the effect of interpolation in computing PTTR values in dosing studies, and how it can distort our understanding of a dosing protocol performance in~\ref{appx: interpolation}.

\subsection{Robustness Check}

We trained and tested our proposed method on a PK/PD model proposed by Hamberg et al.~(2007). As George E. P. Box puts it beautifully, ``all models are wrong, but some are useful.''~\cite{BOX1979201}. It is natural to ask whether the proposed dosing protocol can in fact be useful for real patients. Answering this question requires lengthy and costly dosing trials. Instead, we tested our proposed work by applying it to a different PK/PD model proposed by Hamberg et al.~(2010)~\cite{hamberg2010pharmacometric}. Both models accept the same inputs and estimate the resulting INR as output. However, they have different underlying structures. In the main PK/PD, genotypes of CYP2C9 and VKORC1 are covariates, but in the second PK/PD different alleles are the covariates. For example, if CYP2C9 genotype of ``*1/*1'' is an input to the main PK/PD, the second PK/PD accepts it as two ``*1'' alleles. The main model is a two-compartment model, while the second one is a one-compartment model. Compartments are modeling notions akin to body parts and tissues that contain the medication. The chain of differential equations that translates drug concentration to effect is also different between the two models. Finally, the second PK/PD is based on a much noisier dataset, and consequently variations in the observed INR values are as much as two orders of magnitude higher than in the main PK/PD model.

Table~\ref{tab: pttr_baselines_2nd_pkpd} and Table~\ref{tab: pttr_ours_2nd_pkpd} show the results of applying baseline protocols and our proposed protocols on the second PK/PD model. Comparing baseline results of this PK/PD model and the main PK/PD~(Table~\ref{tab: pttr_base_vs_baselines}) clearly indicates that the two PK/PD models have different behaviors. For example, ``AAA'' protocol achieved 72.4\% PTTR on normal patients when applied on the main PK/PD model. For normal patients of the second PK/PD model, this number is as low as 55.1\%. Also, differences among normal, sensitive, and highly sensitive patients are not as pronounced as when using the main PK/PD model. Our proposed dosing protocols are on par or better than the baseline protocols. For example, our base model ($h=1$ with genotype information) achieves 54.8\% average PTTR for the 10,000 test patients, and the best baseline protocol, ``AAA'', has the average PTTR of 55.1\%. Statistical tests show that $h=3$ is the best performing model (PTTR of 56.7\%), followed by $h=3$ ``no genotype''. The performance ranking using t-test is as follows:

\begin{itemize}
    \item $h=3$
    \item $h=3$ without genotypes
    \item $h=2$ without genotypes \\
          AAA
    \item PGAA
    \item $h=2$
    \item CAA
    \item $h=1$ \\
          PGPGA
    \item $D_{max}^1=5.0 mg$ \\
          PGPGI
    \item $h=1$ without genotypes
    \item $D_{max}^1=5.0 mg$ without genotypes
\end{itemize}

\begin{table}[!ht]
    \caption{Percent Time in Therapeutic Range of baseline protocols on the second PK/PD}
    \label{tab: pttr_baselines_2nd_pkpd}
    \centering
    \begin{tabular}{l l c c c c c}
        \toprule
        protocol    & AAA          & CAA            & PGAA          & PGPGA         & PGPGI \\
        sensitivity \\
        \midrule
        normal      & 55.1\% (0.10)	& 50.6\% (0.14)	& 54.0\% (0.11) & 48.1\% (0.13)	& 43.9\% (0.26) \\
        sensitive   & 55.3\% (0.07)	& 55.6\% (0.07)	& 54.8\% (0.09) & 53.8\% (0.10)	& 54.4\% (0.22) \\
        highly sensitive    & 48.5\% (0.08)	& 49.9\% (0.08)	& 53.7\% (0.10) & 56.0\% (0.07)	& 52.2\% (0.18) \\
        \midrule
        all         & 54.9\% (0.09)	& 52.3\% (0.12)	& 54.3\% (0.10) & 50.4\% (0.13)	& 47.9\% (0.25) \\
        \bottomrule
    \end{tabular}
\end{table}

\begin{table}[!ht]
    \caption{Percent Time in Therapeutic Range of the proposed model on the second PK/PD}
    \label{tab: pttr_ours_2nd_pkpd}
    \centering
    \begin{tabular}{l l c c c c c}
        \toprule
        Parameter      &                   & $h=1$         & $h=2$         & $h=3$         & $D_{max}^1=5mg$ &\\
        genotype availability   & sensitivity \\
        \midrule
        with genotypes      & normal            & 54.8\% (0.09)	& 58.1\% (0.11)	& 58.4\% (0.08) & 50.5\% (0.13) &\\
                            & sensitive         & 42.7\% (0.14)	& 45.9\% (0.11)	& 54.7\% (0.10) & 43.5\% (0.13) &\\
                            & highly sensitive  & 44.3\% (0.11)	& 42.6\% (0.10)	& 48.6\% (0.08) & 42.6\% (0.09) &\\
                            & all               & 50.2\% (0.13)	& 53.3\% (0.13)	& 56.7\% (0.09) & 47.8\% (0.13) &\\
        \midrule[1pt]

        without genotypes   & normal            & 46.0\% (0.09)	& 55.7\% (0.08)	& 55.4\% (0.06) & 45.6\% (0.10) &\\
                            & sensitive         & 46.0\% (0.09)	& 53.5\% (0.11)	& 56.1\% (0.07) & 44.2\% (0.11) &\\
                            & highly sensitive  & 49.6\% (0.09)	& 50.0\% (0.12)	& 54.8\% (0.08) & 49.4\% (0.09) &\\
                            & all               & 46.2\% (0.09)	& 54.7\% (0.09)	& 55.6\% (0.07) & 45.2\% (0.10) &\\
        \bottomrule
    \end{tabular}
\end{table}

\section{Discussion} \label{section: Discussion}
We demonstrated that an RL algorithm coupled with a PK/PD model can produce protocols more effective than common dosing regimens. Testing the proposed work on a different PK/PD model shows the robustness of the proposed work in dealing with highly variable dose-responses and new patient cohorts. Although our protocols are not as effective on new patients (the second PK/PD model) as they are on the original patient cohort (the main PK/PD model), they still manage to perform on par or better than baseline dosing protocols. This provides further evidence that combining mathematical modeling of health phenomena, such as drug dosing, with reinforcement learning is a promising direction for future research. But we need to address the limitations of this approach, possible remedies, and the path toward a usable RL-based dosing protocol.

One such possibility is to customize the PK/PD for each patient. Our work uses a PK/PD model to simulate patients, and the PK/PD captures part of the variability in dose-response as error terms between patients and between measurements in the equations. However, variability in INR measurements of an individual who has a stable lifestyle and eating habits can be much different than an individual with an unstable lifestyle or more varied selection of food. If we can train the RL algorithm on the PK/PD model and further individualize the protocol for each patient based on new observations, then the model can provide a more effective dosing protocol. Lee et al.~(2018) have such an approach in dosing~\cite{Lee2018outcome}. Their work is different from what we discussed in this paper, since they tackle the problem of gestational diabetes, which is a finite horizon model, and use multiobjective mixed integer programming, which plans the full course of treatment rather than proactively updating the treatment as new information arrives. Their novel idea is to build a PK/PD model for each patient based on their measurements and then use that to optimize the intervention for the duration of pregnancy.

The robustness check also reveals how crucial our understanding of the population characteristics and dose-response function is on our ability to create generalizable in-silico dosing protocols. In our work, we simply assumed different patient characteristics are independent and normally distributed. In reality, however, some characteristics of individuals such as genetic makeup impact other characteristics. A more realistic approach is to create virtual patients based on the underlying relationships between the defining characteristics. Ravvaz et al.~(2017), for example, builds a Bayesian network of characteristics using the relationships available in the literature as well as in their dataset~\cite{ravvaz2017personalized}. Not only do these dependencies between characteristics depict a more realistic picture of patient population, they can also eliminate unrealistic or improbable virtual patients which can help the RL methods learn protocols faster and more effectively. We also saw is Section~\ref{section: Results} that rare genotypes negatively impact the performance since the RL algorithm do not see enough samples from those genotypes. Oversampling such genotypes during the training can help with the learning. Oversampling can also help the model avoid converging to protocols that are locally optimal. In our current implementation, we see that PTTR values for normal patients improve faster than for highly sensitive patients. This can impact the model's performance if the model converges to a policy that is optimal for the majority (normal patients), but not for the minorities (highly sensitive patients).

The mathematical model of dose-response is as important as the RL method used for learning. We used PK/PD models in this research. Different PK/PD models exist for warfarin, each of which include a handful of patient characteristics to explain the dose-response. Both of our chosen PK/PD models consider age, CYP2C9, and VKORC1 as the covariates. As a result, other factors that the baseline protocols are including in their decision-making has no effect on the dose-response. A PK/PD with a more comprehensive list of features can better represent the reality of how warfarin affects different patients' INR values. 

One can also consider modeling methods beyond PK/PDs. PK/PD models are great tools to model dose-response interactions from data. However, they are merely modeling tools that do not fully incorporate the physiology of the body. For example, the main PK/PD model we used for training is a two-compartment model~\cite{hamberg2007pk}, and the second model that we used for robustness check is a one-compartment model~\cite{hamberg2010pharmacometric}. These compartments have no direct counterparts in the body, and are decided based on how the resulting behavior matches the observed data. There are two main modeling tools that draw more direct connections to what actually happens in the body, Physiologically-based Pharmacokinetic (PBPK) and Quantitative Systems Pharmacology (QSP) approaches~\cite{pichardo2016from}. The discussion on these approaches is beyond the scope of this paper, but they are known to better reflect the mechanics of how medications affect different markers in the body, and could result in a more generalizable in-silico dosing protocols.

In the end, we need to acknowledge a significant obstacle in the application of our work and similar RL approaches on real patients. The current dosing protocols are all more interpretable than most RL models. Even though protocols like IWPC are non-linear regression models, practitioners can have a better sense of the model and its dosing recommendation than what we offer here. The deep Q-learning method that we utilized in this paper assigns a value to each state-action pair. It is hard, if not impossible, to explain why any given state-action pair has its approximated value. One alternative is to use ``Policy Gradient'' methods that output the optimal action to take. We can then apply post-hoc explainer models to interpret the action (dose) and explain why it is the optimal. It is also possible to substitute the deep neural network core of the method with a more interpretable model, such as rule-based methods or regression methods, to have interpretability incorporated into the dosing protocol itself. Finally, we believe that the \textit{clinician-in-the-loop} approach is a viable path to deployment of our proposed method. In the clinician-in-the-loop approach, the dosing algorithm ``takes into account the action of a clinician as well as her patient’s response when suggesting a new action.''~\cite{nemati2016optimal}. This enables the practitioner to override the dose recommended by our model if she finds the dose unjustifiable.

\section{Conclusion} \label{section: Conclusion}
In this paper, we proposed a deep reinforcement learning-based model for warfarin dosing. Our experiments showed that this approach increases PTTR across the population, and is consistently and significantly superior to available dosing protocols, especially in dealing with highly sensitive patients. We acknowledge that in silico dosing algorithms are only as good as the PK/PD model they employ. The PK/PD model that we chose is one of many such models and does not include factors such as gender, race, and use of other medications that could affect the dosing decision. Usability of the algorithm for practitioners is another issue to take into account. Many dosing algorithms can be summarized into tables with easy to follow \textit{if-this-then-that} rules. These algorithms do not need a software to determine the dose, and since they are relatively simple rule-based models or regression models, physicians and practitioners would trust them easier. We discussed these and limitations of the work in detail in Section~\ref{section: Discussion}. The proposed dosing algorithm can be a viable input for the decision maker as an individualized dosing decision most likely to improve patients' outcome.

\section*{Acknowledgements}
The authors thank Dr. Kourosh Ravvaz for sharing parts of their work.

\bibliography{warfarin_dosing.bib}

\begin{thebibliography}{10}
\expandafter\ifx\csname url\endcsname\relax
  \def\url#1{\texttt{#1}}\fi
\expandafter\ifx\csname urlprefix\endcsname\relax\def\urlprefix{URL }\fi
\expandafter\ifx\csname href\endcsname\relax
  \def\href#1#2{#2} \def\path#1{#1}\fi

\bibitem{krahenbuhl2007drug}
A.~Kr{\"a}henb{\"u}hl-Melcher, R.~Schlienger, M.~Lampert, M.~Haschke, J.~Drewe,
  S.~Kr{\"a}henb{\"u}hl, Drug-related problems in hospitals, {Drug safety}
  30~(5) (2007) 379 -- 407.

\bibitem{maheshwari2018337}
R.~Maheshwari, P.~Sharma, A.~Seth, N.~Taneja, M.~Tekade, R.~K. Tekade,
  \href{http://www.sciencedirect.com/science/article/pii/B9780128144237000101}{{Chapter
  10 - Drug Disposition Considerations in Pharmaceutical Product}}, in: R.~K.
  Tekade (Ed.), {Dosage Form Design Considerations}, {Advances in
  Pharmaceutical Product Development and Research}, Academic Press, 2018, pp.
  337 -- 369.
\newblock \href {https://doi.org/10.1016/B978-0-12-814423-7.00010-1}
  {\path{doi:10.1016/B978-0-12-814423-7.00010-1}}.
\newline\urlprefix\url{http://www.sciencedirect.com/science/article/pii/B9780128144237000101}

\bibitem{chan2017time}
D.~Chan, V.~Ivaturi, J.~Long-Boyle, The time is now: {M}odel-based dosing to
  optimize drug therapy, {International Journal of Pharmacokinetics} 2~(4)
  (2017) 213 -- 215.
\newblock \href {https://doi.org/10.4155/ipk-2017-0011}
  {\path{doi:10.4155/ipk-2017-0011}}.

\bibitem{blix2010drugs}
H.~S. Blix, K.~K. Viktil, T.~A. Moger, A.~Reikvam, {Drugs with Narrow
  Therapeutic Index as Indicators in the Risk Management of Hospitalised
  Patients}, {Pharmacy Practice} 8~(1) (2010) 50.

\bibitem{shaw2015clinical}
K.~Shaw, U.~Amstutz, R.~B. Kim, L.~J. Lesko, J.~Turgeon, V.~Michaud, S.~Hwang,
  S.~Ito, C.~Ross, B.~C. Carleton, et~al., Clinical practice recommendations on
  genetic testing of {CYP2C9} and {VKORC1} variants in warfarin therapy,
  {Therapeutic Drug Monitoring} 37~(4) (2015) 428 -- 436.

\bibitem{moore2014reinforcement}
B.~L. Moore, L.~D. Pyeatt, V.~Kulkarni, P.~Panousis, K.~Padrez, A.~G. Doufas,
  Reinforcement learning for closed-loop propofol anesthesia: {A} study in
  human volunteers, {The Journal of Machine Learning Research} 15~(1) (2014)
  655 -- 696.

\bibitem{hamberg2007pk}
A.-K. Hamberg, M.-L. Dahl, M.~Barban, M.~G. Scordo, M.~Wadelius, V.~Pengo,
  R.~Padrini, E.~N. Jonsson, A {PK--PD} model for predicting the impact of age,
  {CYP2C9}, and {VKORC1} genotype on individualization of warfarin therapy,
  {Clinical Pharmacology \& Therapeutics} 81~(4) (2007) 529 -- 538.
\newblock \href {https://doi.org/10.1038/sj.clpt.6100084}
  {\path{doi:10.1038/sj.clpt.6100084}}.

\bibitem{Alsanosi2014pharmacokinetic}
S.~M.~M. Alsanosi, C.~Skiffington, S.~Padmanabhan,
  \href{www.sciencedirect.com/science/article/pii/B9780123868824000177}{{Chapter
  17 - Pharmacokinetic Pharmacogenomics}}, in: S.~Padmanabhan (Ed.), {Handbook
  of Pharmacogenomics and Stratified Medicine}, {Academic Press}, San Diego,
  2014, pp. 341 -- 364.
\newblock \href {https://doi.org/10.1016/B978-0-12-386882-4.00017-7}
  {\path{doi:10.1016/B978-0-12-386882-4.00017-7}}.
\newline\urlprefix\url{www.sciencedirect.com/science/article/pii/B9780123868824000177}

\bibitem{ho2020trends}
K.~H. Ho, M.~van Hove, G.~Leng, Trends in anticoagulant prescribing: {A} review
  of local policies in {English} primary care, {BMC} health services research
  20 (2020) 1--8.

\bibitem{pirmohamed2006warfarin}
M.~Pirmohamed, Warfarin: {A}lmost 60 years old and still causing problems,
  {British Journal of Clinical Pharmacology} 62~(5) (2006) 509.

\bibitem{fusaro2013systems}
V.~A. Fusaro, P.~Patil, C.-L. Chi, C.~F. Contant, P.~J. Tonellato, A systems
  approach to designing effective clinical trials using simulations,
  {Circulation} 127~(4) (2013) 517 -- 526.

\bibitem{gage2008use}
B.~F. Gage, C.~Eby, J.~A. Johnson, E.~Deych, M.~J. Rieder, P.~M. Ridker, P.~E.
  Milligan, G.~Grice, P.~Lenzini, A.~E. Rettie, A.~E. Rettie, C.~L. Aquilante,
  Use of pharmacogenetic and clinical factors to predict the therapeutic dose
  of warfarin, {Clinical Pharmacology \& Therapeutics} 84~(3) (2008) 326 --
  331.
\newblock \href {https://doi.org/10.1038/clpt.2008.10}
  {\path{doi:10.1038/clpt.2008.10}}.

\bibitem{international2009estimation}
{International Warfarin Pharmacogenetics Consortium}, Estimation of the
  warfarin dose with clinical and pharmacogenetic data, {New England Journal of
  Medicine} 360~(8) (2009) 753 -- 764.

\bibitem{anderson2007randomized}
J.~L. Anderson, B.~D. Horne, S.~M. Stevens, A.~S. Grove, S.~Barton, Z.~P.
  Nicholas, S.~F.~S. Kahn, H.~T. May, K.~M. Samuelson, J.~B. Muhlestein, J.~F.
  Carlquist, {Randomized Trial of Genotype-Guided Versus Standard Warfarin
  Dosing in Patients Initiating Oral Anticoagulation}, {Circulation} 116~(22)
  (2007) 2563 -- 2570.
\newblock \href {https://doi.org/10.1161/CIRCULATIONAHA.107.737312}
  {\path{doi:10.1161/CIRCULATIONAHA.107.737312}}.

\bibitem{avery2011a}
P.~J. Avery, A.~Jorgensen, A.~K. Hamberg, M.~Wadelius, M.~Pirmohamed,
  F.~Kamali, E.-P.~S. Group, A proposal for an individualized
  pharmacogenetics-based warfarin initiation dose regimen for patients
  commencing anticoagulation therapy, {Clinical Pharmacology and Therapeutics}
  90~(5) (2011) 701 -- 706.
\newblock \href {https://doi.org/10.1038/clpt.2011.186}
  {\path{doi:10.1038/clpt.2011.186}}.

\bibitem{johnson2017clinical}
J.~A. Johnson, K.~E. Caudle, L.~Gong, M.~Whirl-Carrillo, C.~M. Stein, S.~A.
  Scott, M.~T. Lee, B.~F. Gage, S.~E. Kimmel, M.~A. Perera, et~al., {Clinical
  Pharmacogenetics Implementation Consortium (CPIC)} guideline for
  pharmacogenetics-guided warfarin dosing: 2017 update, {Clinical Pharmacology
  \& Therapeutics} 102~(3) (2017) 397 -- 404.

\bibitem{lenzini2010integration}
P.~Lenzini, M.~Wadelius, S.~Kimmel, J.~L. Anderson, A.~L. Jorgensen,
  M.~Pirmohamed, M.~D. Caldwell, N.~Limdi, J.~K. Burmester, M.~B. Dowd, et~al.,
  Integration of genetic, clinical, and {INR} data to refine warfarin dosing,
  {Clinical Pharmacology \& Therapeutics} 87~(5) (2010) 572 -- 578.

\bibitem{pirmohamed2013randomized}
M.~Pirmohamed, G.~Burnside, N.~Eriksson, A.~L. Jorgensen, C.~H. Toh,
  T.~Nicholson, P.~Kesteven, C.~Christersson, B.~Wahlstr{\"o}m, C.~Stafberg,
  et~al., A randomized trial of genotype-guided dosing of warfarin, {New
  England Journal of Medicine} 369 (2013) 2294 -- 2303.

\bibitem{yu2019reinforcement}
C.~Yu, J.~Liu, S.~Nemati, Reinforcement learning in healthcare: {A} survey,
  {arXiv preprint arXiv:1908.08796} (2019).

\bibitem{shortreed2011informing}
S.~M. Shortreed, E.~Laber, D.~J. Lizotte, T.~S. Stroup, J.~Pineau, S.~A.
  Murphy, Informing sequential clinical decision-making through reinforcement
  learning: {An} empirical study, {Machine learning} 84~(1-2) (2011) 109 --
  136.

\bibitem{levy2019applications}
A.~E. Levy, M.~Biswas, R.~Weber, K.~Tarakji, M.~Chung, P.~A. Noseworthy,
  C.~Newton-Cheh, M.~A. Rosenberg, Applications of machine learning in decision
  analysis for dose management for dofetilide, {PloS} one 14~(12) (2019).

\bibitem{nemati2016optimal}
S.~Nemati, M.~M. Ghassemi, G.~D. Clifford, Optimal medication dosing from
  suboptimal clinical examples: {A} deep reinforcement learning approach, in:
  2016 38th Annual International Conference of the {IEEE Engineering in
  Medicine and Biology Society (EMBC)}, IEEE, 2016, pp. 2978 -- 2981.

\bibitem{padmanabhan2017reinforcement}
R.~Padmanabhan, N.~Meskin, W.~M. Haddad, Reinforcement learning-based control
  of drug dosing for cancer chemotherapy treatment, {Mathematical biosciences}
  293 (2017) 11 -- 20.

\bibitem{humphrey2017using}
K.~Humphrey, \href{http://hdl.handle.net/10150/625341}{Using reinforcement
  learning to personalize dosing strategies in a simulated cancer trial with
  high dimensional data}, Master's thesis, The University of Arizona (4 2017).
\newline\urlprefix\url{http://hdl.handle.net/10150/625341}

\bibitem{hamberg2010pharmacometric}
A.-K. Hamberg, M.~Wadelius, J.-D. Lindh, M.-L. Dahl, R.~Padrini, P.~Deloukas,
  A.~Rane, E.~N. Jonsson, A pharmacometric model describing the relationship
  between warfarin dose and {INR} response with respect to variations in
  {CYP2C9}, {VKORC1}, and age, {Clinical Pharmacology \& Therapeutics} 87~(6)
  (2010) 727 -- 734.
\newblock \href {https://doi.org/10.1038/clpt.2010.37}
  {\path{doi:10.1038/clpt.2010.37}}.

\bibitem{ravvaz2017personalized}
K.~Ravvaz, J.~A. Weissert, C.~T. Ruff, C.-L. Chi, P.~J. Tonellato, Personalized
  anticoagulation: {O}ptimizing warfarin management using genetics and
  simulated clinical trials, {Circulation: Cardiovascular Genetics} 10~(6)
  (2017) e001804.
\newblock \href {https://doi.org/10.1161/CIRCGENETICS.117.001804}
  {\path{doi:10.1161/CIRCGENETICS.117.001804}}.

\bibitem{UCSD-tablet}
\href{https://health.ucsd.edu/specialties/anticoagulation/providers/warfarin/pages/tablet-identification.aspx}{Warfarin
  tablet identification} (2021).
\newline\urlprefix\url{https://health.ucsd.edu/specialties/anticoagulation/providers/warfarin/pages/tablet-identification.aspx}

\bibitem{hirsh2003american}
J.~Hirsh, V.~Fuster, J.~Ansell, J.~L. Halperin, American heart
  association/american college of cardiology foundation guide to warfarin
  therapy, {Journal of the American College of Cardiology} 41~(9) (2003) 1633
  -- 1652.

\bibitem{biedermann2017optimization}
J.~S. Biedermann, \href{http://hdl.handle.net/1765/102849}{{Optimization of
  Monitoring and Management of Anticoagulant Therapy}}, Ph.D. thesis,
  Department of Hematology (11 2017).
\newline\urlprefix\url{http://hdl.handle.net/1765/102849}

\bibitem{rosendaal1993method}
F.~R. Rosendaal, S.~C. Cannegieter, F.~J.~M. Van~der Meer, E.~Briet, A method
  to determine the optimal intensity of oral anticoagulant therapy, {Thrombosis
  and Haemostasis} 70~(3) (1993) 236 -- 239.

\bibitem{powell2020reinforcement}
W.~Powell, {Reinforcement Learning and Stochastic Optimization: A Unified
  Framework for Sequential Decisions}, Princeton NJ (2020).

\bibitem{hornik1991approximation}
K.~Hornik, Approximation capabilities of multilayer feedforward networks,
  Neural networks 4~(2) (1991) 251 -- 257.

\bibitem{mnih2015human}
V.~Mnih, K.~Kavukcuoglu, D.~Silver, A.~A. Rusu, J.~Veness, M.~G. Bellemare,
  A.~Graves, M.~Riedmiller, A.~K. Fidjeland, G.~Ostrovski, et~al., Human-level
  control through deep reinforcement learning, {Nature} 518~(7540) (2015) 529
  -- 533.

\bibitem{sutton2018reinforcement}
R.~S. Sutton, A.~G. Barto, {Reinforcement Learning: An Introduction}, {MIT
  press}, 2018.

\bibitem{BOX1979201}
G.~Box,
  \href{https://www.sciencedirect.com/science/article/pii/B9780124381506500182}{Robustness
  in the strategy of scientific model building}, in: R.~L. LAUNER, G.~N.
  WILKINSON (Eds.), Robustness in Statistics, Academic Press, 1979, pp.
  201--236.
\newblock \href
  {https://doi.org/https://doi.org/10.1016/B978-0-12-438150-6.50018-2}
  {\path{doi:https://doi.org/10.1016/B978-0-12-438150-6.50018-2}}.
\newline\urlprefix\url{https://www.sciencedirect.com/science/article/pii/B9780124381506500182}

\bibitem{Lee2018outcome}
E.~K. Lee, X.~Wei, F.~Baker-Witt, M.~D. Wright, A.~Quarshie, Outcome-driven
  personalized treatment design for managing diabetes, {INFORMS Journal on
  Applied Analytics} 48~(5) (2018) 422--435.

\bibitem{pichardo2016from}
C.~Pichardo-Almarza, V.~Diaz-Zuccarini, From {PK/PD} to {QSP}: {U}nderstanding
  the dynamic effect of cholesterol-lowering drugs on atherosclerosis
  progression and stratified medicine, Current pharmaceutical design 22~(46)
  (2016) 6903 -- 6910.
\newblock \href {https://doi.org/10.2174/1381612822666160905095402}
  {\path{doi:10.2174/1381612822666160905095402}}.

\end{thebibliography}

\clearpage
\appendix
\section{Covariates of the four main pharmacogenetic dosing protocols} \label{appx: main_protocols_variables}
The four main warfarin dosing algorithms that incorporate genetic information are: Gage~et~al.~\cite{gage2008use}, IWPC~\cite{international2009estimation}, Lenzini~et~al.~\cite{lenzini2010integration}, and EU-PACT~\cite{pirmohamed2013randomized}.Table~\ref{tab: variables_of_major_dosing_protocols} summarizes the covariates that each of these algorithms consider in their dose prediction.

\begin{table}[!ht]
    \caption{Covariates of the four main pharmacogenetic warfarin dosing protocols}
    \label{tab: variables_of_major_dosing_protocols}
    \centering
    \begin{threeparttable}
        \begin{tabular}{l c c c c}
            \toprule
             Variable                      	        & IWPC                  & EU-PACT           & Gage et al.       & Lenzini \\
             \midrule 
             Age                                    & \checkmark\tnote{*}   & \checkmark        & \checkmark\tnote{*}   & \checkmark \\
             Height                                 & \checkmark            & \checkmark        &                   &  \\
             Weight                                 & \checkmark            & \checkmark        &                   &  \\
             BSA                                    &                       &                   & \checkmark        & \checkmark \\
             CYP2C9                                 &                       &                   &                   &  \\
                                                    & *1/*2                 & *1/*2             & *2\tnote{\dag}    & *2\tnote{\dag}  \\
                                                    & *1/*3                 & *1/*3             & *3\tnote{\dag}    & *2\tnote{\dag}  \\
                                                    & *2/*2                 & *2/*2             &                   &   \\
                                                    & *2/*3                 & *2/*3             &                   &   \\
                                                    & *3/*3                 & *3/*3             &                   &   \\
                                                    & Unknown               &                   &                   &   \\
             VKORC1                                 &                       &                   & \checkmark\tnote{\ddag}   & \checkmark\tnote{\ddag}  \\
             \hspace{5mm}A/G                        & \checkmark            & \checkmark        &                   & \\
             \hspace{5mm}A/A                        & \checkmark            & \checkmark        &                   &   \\
             \hspace{5mm}Unknown                    & \checkmark            &                   &                   &   \\
             Race                                   &                       &                   &                   &   \\
             \hspace{5mm}Black or African American  & \checkmark            &                   & \checkmark        &  \checkmark \\
             \hspace{5mm}Asian                      & \checkmark            &                   &                   &   \\
             \hspace{5mm}Missing or Mixed race      & \checkmark            &                   &                   &   \\
             Enzyme inducer                         & \checkmark            &                   &                   &   \\
             Amiodarone                             & \checkmark            & \checkmark        & \checkmark        & \checkmark  \\
             Target INR                             &                       &                   & \checkmark        & \checkmark  \\
             Smoking                                &                       &                   & \checkmark        &   \\
             Deep venous thrombosis (DVT) or        &                       &                   &                   &  \\
             pulmonary embolism (PE)                &                       &                   & \checkmark        &  \\
             INR                                    &                       &                   &                   & \checkmark\tnote{\S} \\
             Stroke                                 &                       &                   &                   & \checkmark  \\
             Diabetes                               &                       &                   &                   & \checkmark  \\
             Fluvastatin                            &                       &                   &                   & \checkmark  \\
             Previous doses                         &                       &                   &                   & \checkmark\tnote{$||$} \\
             \bottomrule
        \end{tabular}
        \begin{tablenotes}
            \item[*] {per decade}
            \item[\dag] {0 if absent, 1 if heterozygous, 2 if homozygous}
            \item[\ddag] {0 for G/G, 1 for A/G, 2 for A/A}
            \item[\S] {Natural logarithm}
            \item[$||$] {Doses of two, three and four days ago}
        \end{tablenotes}
    \end{threeparttable}
\end{table}

\clearpage
\section{Study of individual patients} \label{appx: individuals}
In this appendix, we study two examples of patients for whom our proposed method fails while at least one baseline does an acceptable job. We also show examples that the baselines fail and our model performs well.

\begin{figure}[h]
    \centering
    \begin{subfigure}[b]{\textwidth}
        \centering
        \includegraphics[width=\textwidth]{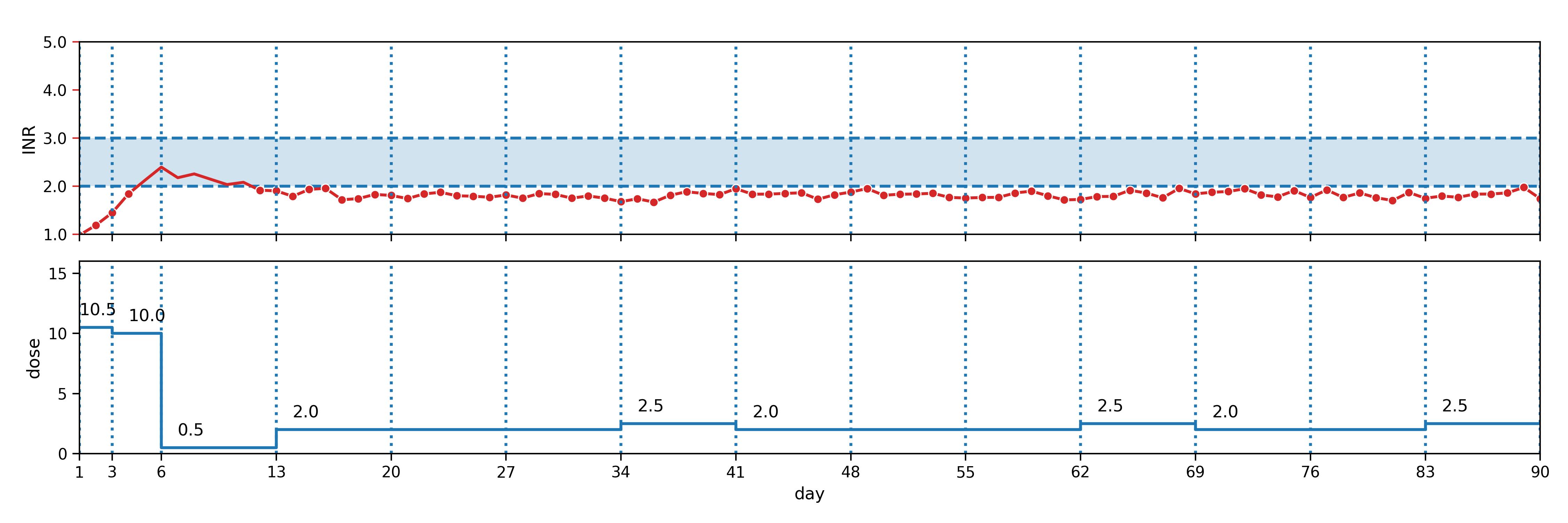}
        \caption{Base model}
        \label{fig: patient_1017790_base}
    \end{subfigure}

    \begin{subfigure}[b]{\textwidth}
        \centering
        \includegraphics[width=\textwidth]{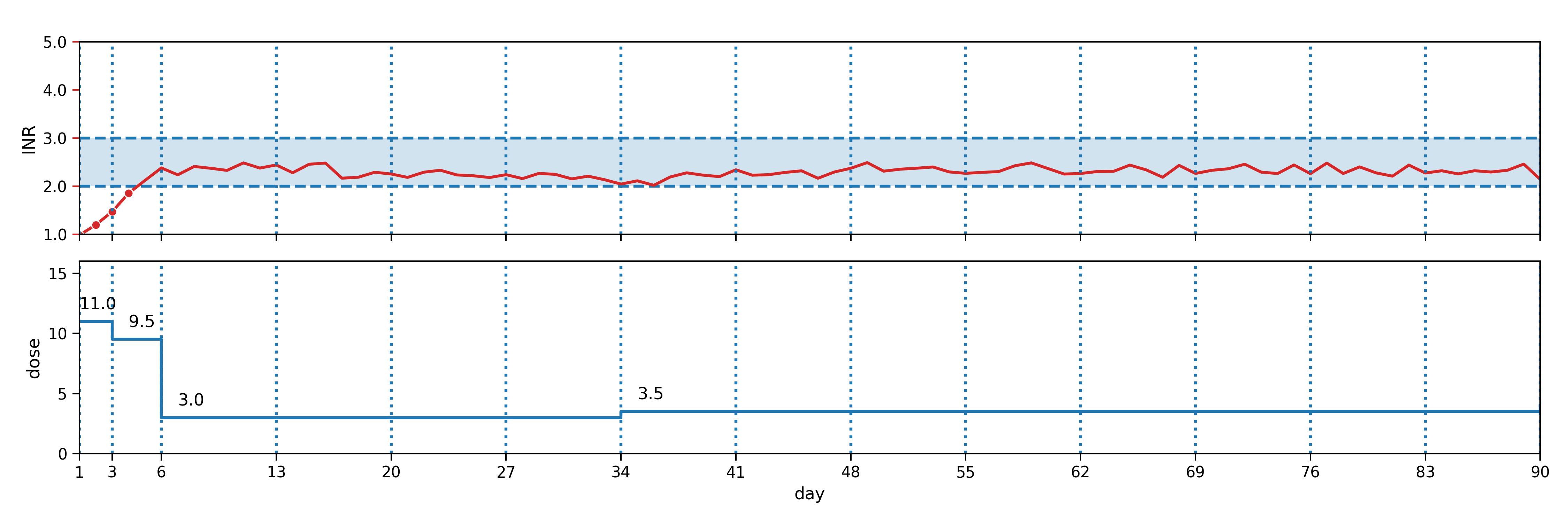}
        \caption{Improved performance in ``without genotype'' model}
        \label{fig: patient_1017790_no}
    \end{subfigure}
    \caption{Patient 1017790: a sensitive patient with low PTTR under the proposed protocol}
    \label{fig: patient_1017790}
\end{figure}

Consider Patient 1017790 who is a sensitive patient with the biggest drop in performance among all patients by 81.1 percentage points. Figure~\ref{fig: patient_1017790_base} plots dose and INR values for the 90-day dosing trial. The red dots on the INR plot highlight out-of-range INR values and dashed vertical lines indicate the decision points. The prescribed dose at each dosing interval is annotated on the dose plot. Initially, the model acts quickly and moves the patient's INR to the therapeutic range. But it fails to identify the necessary dose later on. On the contrary, the ``without genotype'' variant effectively manages the INR and keeps the patient in the range with only three dose changes during the trial~\ref{fig: patient_1017790_no}.

\begin{figure}
    \centering
    \begin{subfigure}[b]{\textwidth}
        \centering
        \includegraphics[width=\textwidth]{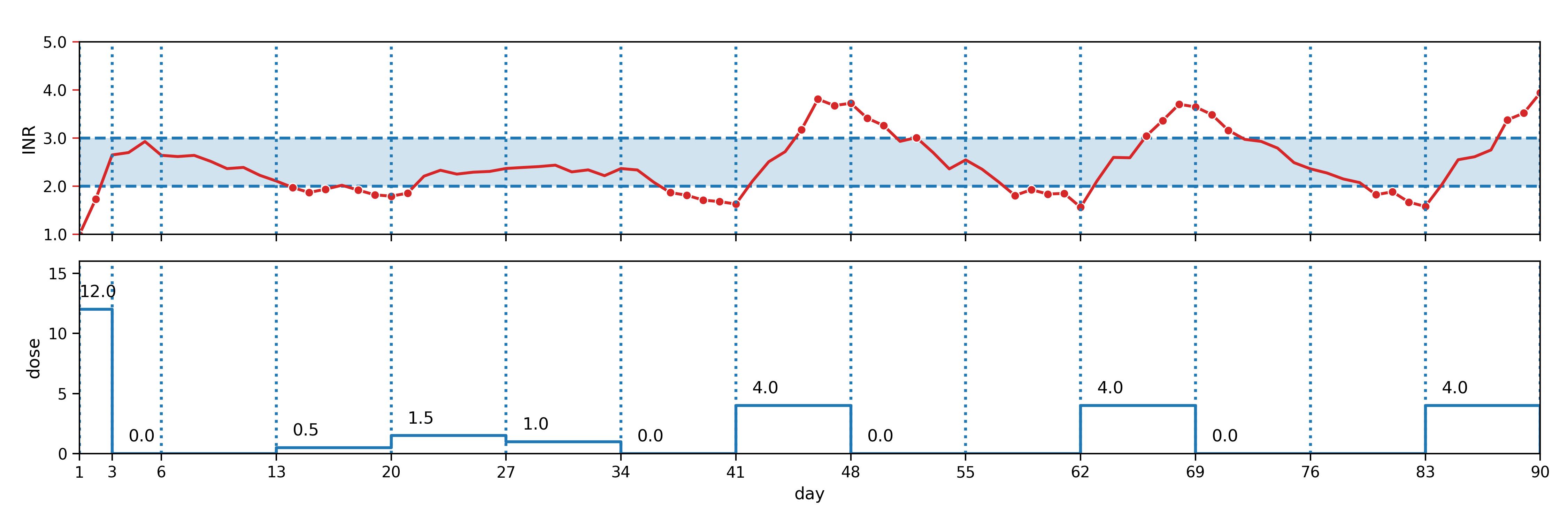}
        \caption{Base model}
        \label{fig: patient_1015572_base}
    \end{subfigure}

    \begin{subfigure}[b]{\textwidth}
        \centering
        \includegraphics[width=\textwidth]{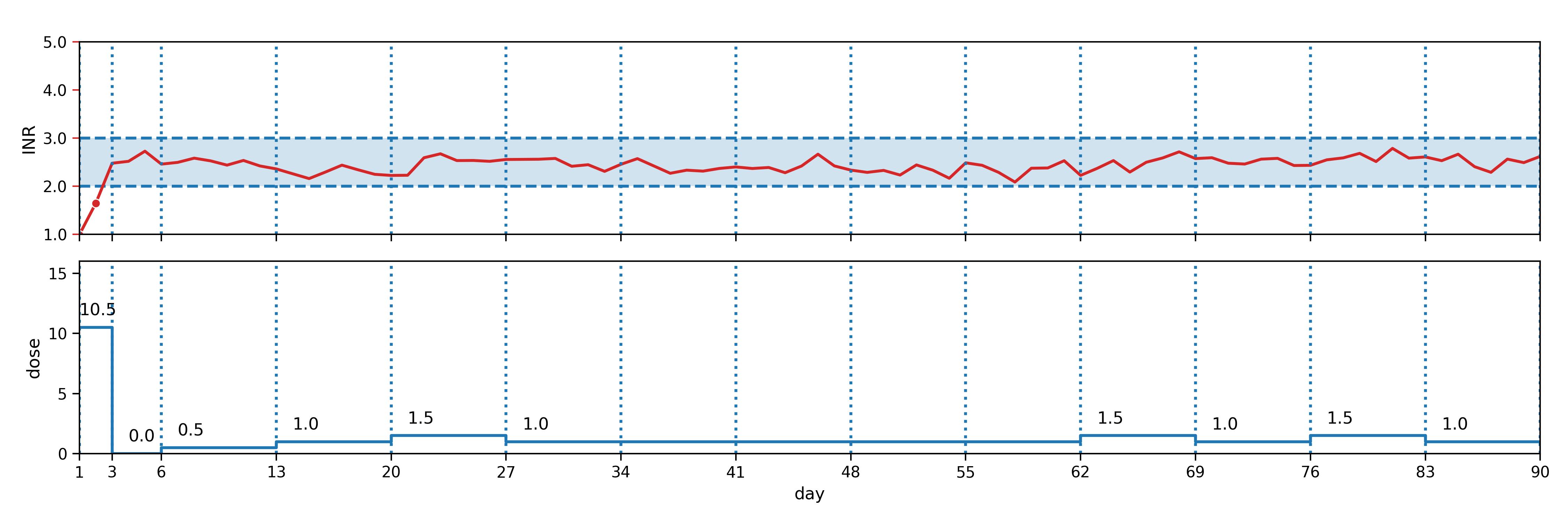}
        \caption{Improved performance in ``without genotype'' model}
        \label{fig: patient_1015572_no}
    \end{subfigure}

    \caption{Patient 1015572: a highly sensitive patient with low PTTR under the proposed protocol}
    \label{fig: patient_1015572}
\end{figure}

The second example is the worst performing highly sensitive patient under our base model. As Figure~\ref{fig: patient_1015572_base} depicts, Patient 1015572 requires a small amount of warfarin, and the model has a hard time adjusting the dose. The ``without genotype'' variant is doing a better job, and puts the patient in the therapeutic state in day two and controls the INR effectively for the rest of the dosing trial~(Figure~\ref{fig: patient_1015572_no}).

\begin{figure}
    \centering
    \begin{subfigure}[b]{\textwidth}
        \centering
        \includegraphics[width=\textwidth]{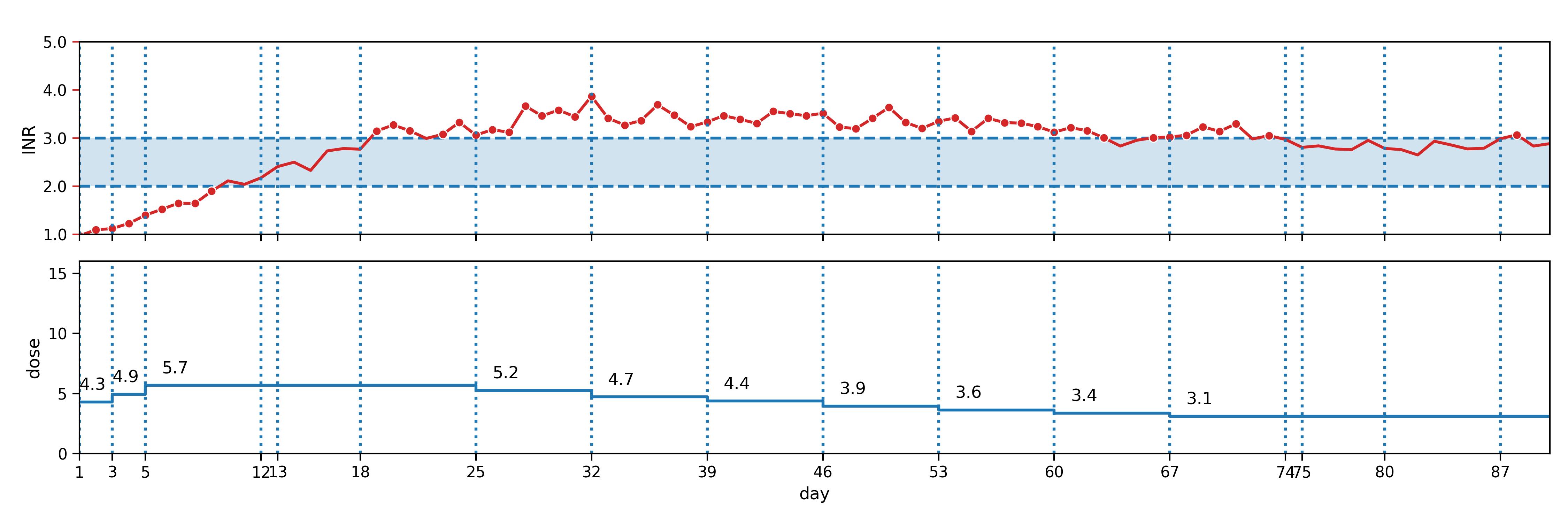}
        \caption{Patient 1017979 - Normal - CAA Protocol}
        \label{fig: patient_1017979}
    \end{subfigure}

    \begin{subfigure}[b]{\textwidth}
        \centering
        \includegraphics[width=\textwidth]{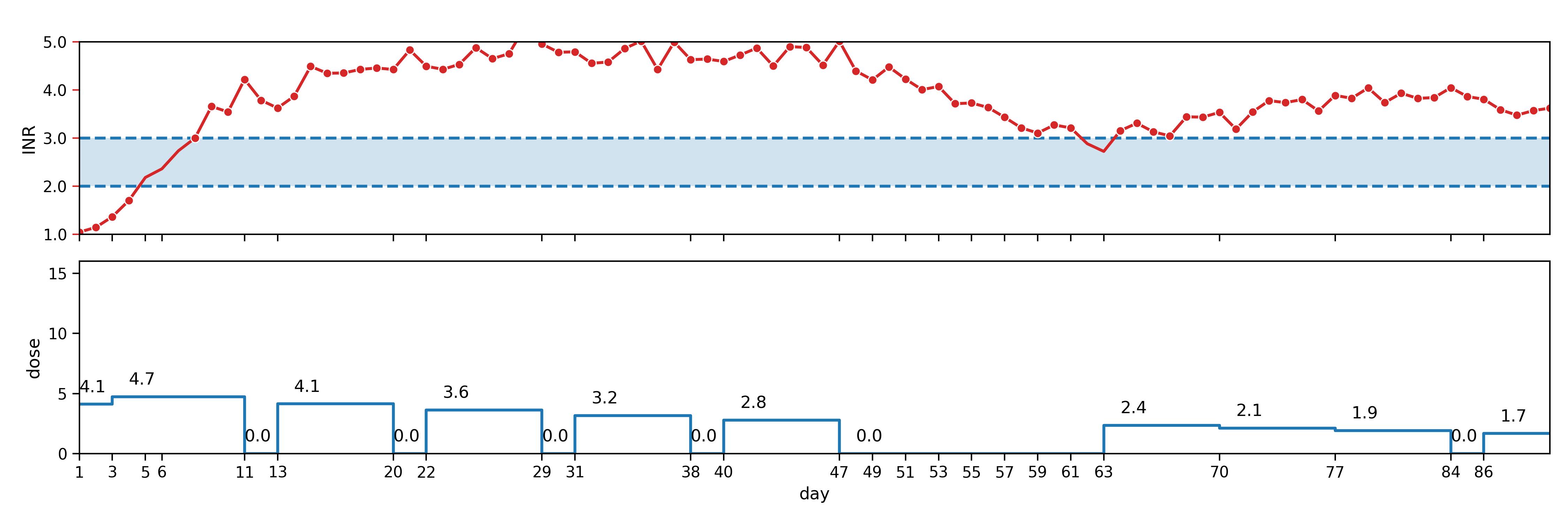}
        \caption{Patient 1014466 - Sensitive - CAA Protocol}
        \label{fig: patient_1014466}
    \end{subfigure}

    \begin{subfigure}[b]{\textwidth}
        \centering
        \includegraphics[width=\textwidth]{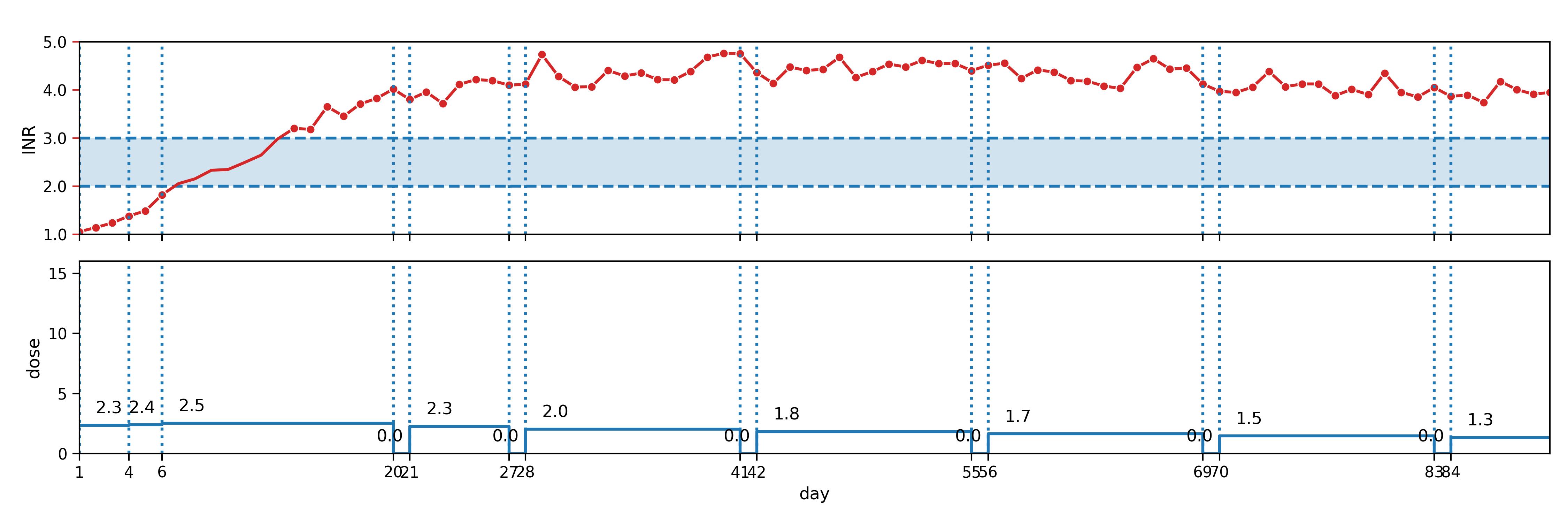}
        \caption{Patient 1019157 - Highly sensitive - PGPGI Protocol}
        \label{fig: patient_1019157}
    \end{subfigure}
    \caption{Example of patients with low PTTR in their best baseline protocol}
    \label{fig: worst_baselines}
\end{figure}

On the other side of the performance gap spectrum are patients for whom baselines are not as effective. Slow change in dose is the main cause of problem in these cases. Figure~\ref{fig: worst_baselines} shows three cases under baseline protocols. In the normal case (Figure~\ref{fig: patient_1017979}), CAA protocol fails to reduce the dose fast enough to return the patient into the therapeutic range. For both sensitive (Figure~\ref{fig: patient_1014466}) and highly sensitive (Figure~\ref{fig: patient_1019157}) cases, the protocol starts with a safe low dose, but the patient passes the therapeutic range quickly and gradual decrease in dose does not fix the problem. Our base model achieves PTTR values of $87.8\%$, $96.7\%$, and $97.8\%$ for these patients, respectively.

\clearpage
\section{Effect of INR interpolation on PTTR} \label{appx: interpolation}

\begin{figure}[h]
    \centering
    \includegraphics[width=\textwidth]{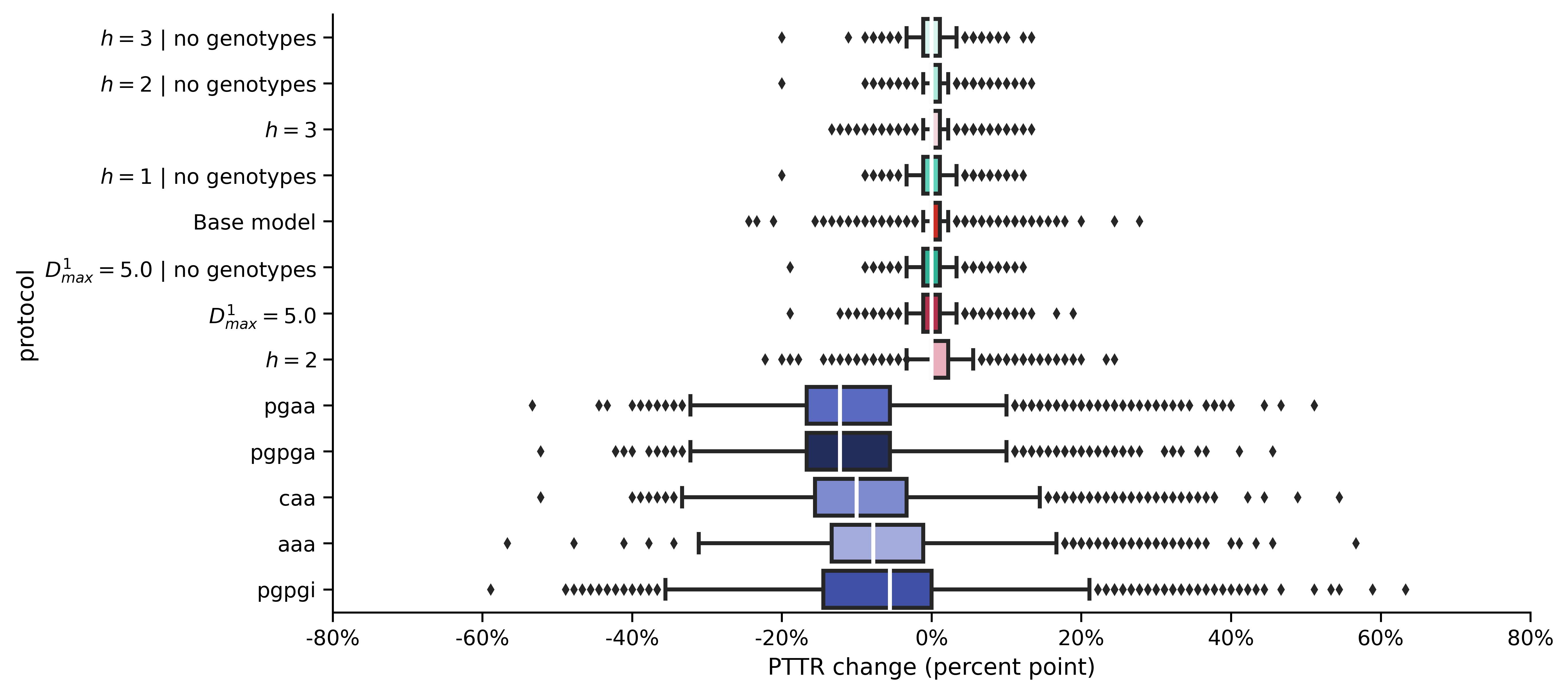}
    \caption{The amount of change in PTTR (in percentage points) if interpolated values are used}
    \label{fig: daily_vs_interpolated_pttr}
\end{figure}

\begin{figure}[h]
    \centering
    \includegraphics[width=\textwidth]{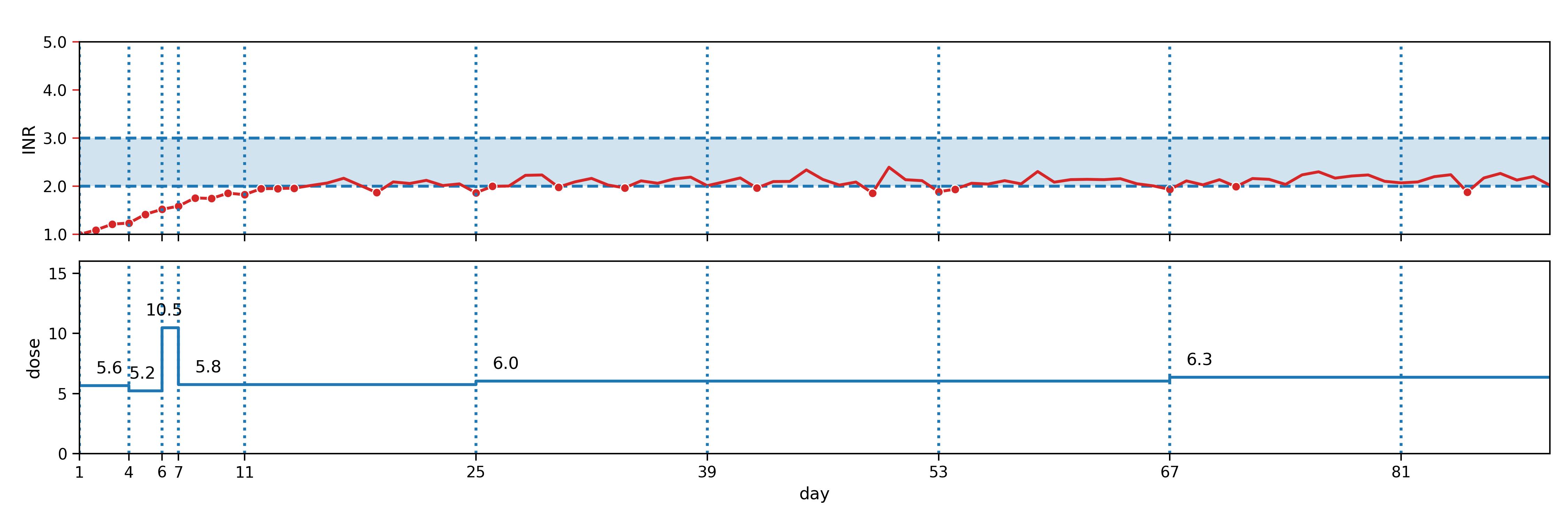}
    \caption{Patient 1016850: Most observed INR values are out of range}
    \label{fig: patient_1016850_pgpgi}
\end{figure}

In this appendix, we would like to briefly discuss the importance of the metrics used in this line of research. None of the baseline dosing protocols used in this research rely on daily measurements. Although the reward function in the training of our models uses daily INR values, the models receive no such information when making dosing decisions. Therefore, our models do not rely on daily measurements either. However, we used daily measurements to compute PTTR values and compare protocols. As discussed in Section~\ref{subsection: Problem Description}, daily measurement of INR is not practical, and it is a common practice to interpolate unobserved measurements. Rosendaal interpolation (\cite{rosendaal1993method}) is one of these methods that linearly interpolates values, and is being used to compare warfarin protocols in~\cite{ravvaz2017personalized}. In this research, we used daily INR values, as they were available by solving the PK/PD model. To see if interpolation and daily measurements provide comparable means of analyzing protocols, we compared PTTR values computed using these two techniques. Figure~\ref{fig: daily_vs_interpolated_pttr} shows how interpolation distorts the outcome. The vertical axis shows all models in this study ranked by their performance. The horizontal axis shows the percentage point change in PTTR value of patients if we use interpolated INR values instead of daily measurements. For our models, interpolation marginally increases the PTTR value since we test more frequently and also keep patients in the range more effectively. On the contrary, baseline PTTRs using interpolation are much smaller than the actual PTTR values. The median change for all baseline protocols are negative. One possible explanation is that for baseline protocols, measurements can be many days apart. If on both ends of an interval, the measured INR is out-of-range, then the interpolation counts all INR values of that interval as out-of-range. For example, Patient 1016850 is a normal patient whose PTTR under PGPGI protocol is $71.1\%$~(Figure~\ref{fig: patient_1016850_pgpgi}). If we want to use interpolated values, the patient will be assumed to be out of range for the interval of day 11 to day 25 and day 53 to day 67. The interpolated PTTR for this patient will be $12.2\%$. Therefore, for the interpolated PTTR, only days 62 to 90 are in-range. Such patterns distort our metric and affects our understanding of the protocols and their effectiveness.

\end{document}